\documentclass[11pt]{article}

\usepackage[final]{acl}

\usepackage{times}
\usepackage{latexsym}
\usepackage{amsfonts}
\usepackage{amsmath}
\usepackage{booktabs}
\usepackage{xcolor}

\usepackage[T1]{fontenc}
\usepackage[utf8]{inputenc}

\usepackage{microtype}

\usepackage{inconsolata}

\usepackage{graphicx}

\usepackage{algorithm}
\usepackage{algpseudocode}

\usepackage{xcolor}
\usepackage{soul}

\usepackage{multirow}

\newcommand{\gr}[1]{\textcolor{gray}{#1}}

\title{C2: Scalable Rubric-Augmented Reward Modeling from Binary Preferences}

\author{
  Akira Kawabata\textsuperscript{1,2,3}\thanks{Work done while at The Asahi Shimbun Company.} \quad Saku Sugawara\textsuperscript{1,2,4} \\
  \textsuperscript{1}The Graduate University for Advanced Studies (SOKENDAI) \\
  \textsuperscript{2}National Institute of Informatics\quad
  \textsuperscript{3}The Asahi Shimbun Company \\
  \textsuperscript{4}The University of Tokyo \\
  \texttt{\{akira, saku\}@nii.ac.jp}
}

\begin{document}
\maketitle
\begin{abstract}

Rubric-augmented verification guides reward models with explicit evaluation criteria, yielding more reliable judgments than single-model verification.
However, most existing methods require costly rubric annotations, limiting scalability.
Moreover, we find that rubric generation is vulnerable to a failure of cooperation; low-quality rubrics actively mislead reward models rather than help. 
Inspired by the principle of cooperative communication, we propose Cooperative yet Critical reward modeling (C2), a framework that significantly improves reward model judgments by having the reward model critically collaborate with a rubric generator trained solely from binary preferences.
In C2, we synthesize helpful and misleading rubric pairs by measuring how each rubric shifts the reward model toward or away from the correct preference. 
Using these contrastive pairs, we train a cooperative rubric generator to propose helpful rubrics, and a critical verifier to assess rubric validity before making its judgment, following only rubrics it deems helpful at inference time. 
C2 outperforms reasoning reward models trained on the same binary preferences, with gains of up to 6.5 points on RM-Bench and 6.0 points length-controlled win rate on AlpacaEval 2.0. 
Without external rubric annotations, C2 enables an 8B reward model to match performance achieved with rubrics from a 4$\times$ larger model.
Overall, our work demonstrates that eliciting deliberate cooperation in rubric-augmented verification makes reward models more trustworthy in a scalable way.\footnote{Our code is available at \url{https://github.com/asahi-research/C2}.}

\end{abstract}

\section{Introduction}

Aligning large language models with human values is critical for their reliable deployment \cite{ouyang2022training}. 
Reinforcement Learning from Human Feedback (RLHF) provides a principled framework for this alignment \cite{NIPS2017_rlhf}. 
Central to RLHF are verifiers that act as scalable proxies for human judgments, trained via reward modeling on binary preferences, i.e., pairwise judgments indicating the better output \cite{Stiennon2020learning, bai2022constitutional}.
However, providing robust verification remains challenging in domains where evaluation criteria are implicit and subjective, such as creative writing and instruction following \cite{eisenstein2024helping, ying2025correctness}.
Rubric-augmented verification addresses this by guiding verifiers with rubrics decomposing evaluation into tractable sub-questions, yielding more reliable judgments than a single verifier \cite{viswanathan2025checklists}.

\begin{figure}[t]
    \includegraphics[width=\linewidth]{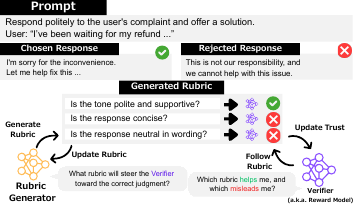}
    \caption{We frame rubric generation and rubric-grounded verification as cooperative yet critical communication: the generator cooperatively explores rubrics to guide the verifier toward correct judgments, and the verifier critically assesses which rubrics to follow based on their outcomes.}
    \label{fig:fig_1}
\end{figure}

Rubric-augmented verification is promising, but most methods rely on rubrics from human annotators or proprietary models \cite{chen2025rmr1reward, gunjal2025rubricsrewards}. 
Unlike conventional reward modeling, which utilizes widely available binary preferences \cite{wang-etal-2025-helpsteer3, liu2025skyworkrewardv2}, this reliance on fine-grained annotations incurs substantial costs and limits the reuse of existing preference corpora. 
Consequently, those rubric-based methods are less viable as scalable alternatives to the current methods.
Given these limitations, a natural alternative is to use self-generated rubrics.
However, our experiments indicate that such self-generated rubrics often vary in quality and, on average, do not enable verifiers to make more accurate judgments.
Looking more closely, we find that rubric quality decisively affects verifier judgments.
High-quality rubrics that are discriminative and consistent with the question's intent enable verifiers to make far more accurate judgments. 
By contrast, vague or misaligned rubrics can severely distort verifier reasoning, pushing it toward incorrect judgments even when the verifier would have judged correctly on its own. 
This amounts to a \emph{failure of cooperation} between the rubric generator and the verifier, where the generated rubric actively misleads rather than helps. 
We therefore ask: Can we design a rubric-augmented verification that is both scalable and robust to such failures, using only binary preferences as supervision?

To answer this, we draw inspiration from theories of cooperative communication \cite{Grice1975-GRILAC-6, Sperber1986-yu}.
Human communication succeeds not because speakers are always reliable, but because both sides adapt: speakers learn which signals help listeners, and listeners learn which speakers to trust \cite{Clark1991-CLAGIC, Sperber2010-SPEEV}. 
We hypothesize that the same dynamic governs rubric generation and rubric-based verification (Figure~\ref{fig:fig_1}): the generator learns which rubrics help; the verifier learns which to trust.

Based on this insight, we propose Cooperative yet Critical reward modeling (C2), a framework jointly training a rubric generator and a rubric-augmented verifier.
The core idea is to synthesize contrastive rubric pairs based on whether each rubric helps or misleads the verifier, and use them to supervise both the generator and verifier.
The cooperative generator is trained via Direct Preference Optimization (DPO; \citealp{rafailov2023direct}) on these contrastive pairs to produce helpful rubrics. 
The critical verifier is trained via Group Relative Policy Optimization (GRPO; \citealp{shao2024deepseek}) to reason about which response is better and whether to trust the rubric.
At inference, the verifier follows rubrics it deems helpful and reverts to rubric-free evaluation otherwise.
C2 thus enables scalable rubric-augmented verification from binary preferences alone by training the rubric generator and verifier to cooperate critically.

In summary, our contributions are as follows:
\begin{itemize}
\item We empirically characterize the two-sided nature of self-generated rubrics: most have negligible impact, but high-quality ones substantially improve accuracy whereas low-quality ones actively hurt it.

\item We propose C2, a framework that realizes rubric-augmented verification without external rubric annotations. C2 synthesizes helpful and misleading rubrics from binary preferences to train a cooperative rubric generator and a critical verifier, with selective inference.

\item Experiments on two base models confirm C2 outperforms reasoning reward models trained with GRPO, a method central to recent state-of-the-art verifiers, in both preference prediction (+6.5 points on RM-Bench) and RLHF (+6.0 points LC win rate on AlpacaEval).

\end{itemize}

\section{Related Work}
\label{sec_related_works}

\subsection{Reward Models}
Reward models (RMs) serve as learned proxies for human preferences \cite{bai2022traininghelpful, kawabata-sugawara-2024-rationale, wang2025helpsteerpreference}.
In RLHF, they provide the reward signal for policy optimization.
At inference, RMs rank candidates, trading compute for quality \cite{cobbe2021trainingverifier, snell2025scaling}.
However, scalar RMs are sensitive to superficial features and generalize poorly out-of-domain \cite{bukharin2025adversarial, liu2025rmbench, wu-etal-2025-rewordbench}.
To address this, state-of-the-art verifiers such as J1 and Think-RM frame preference prediction as a reasoning task optimized with GRPO \cite{whitehouse2025j1, hong2025thinkrm, guo2025reward}, achieving stronger generalization. 
Our work builds on this reasoning-based approach but moves beyond single-verifier by jointly training rubric generator and verifier from preference data.

\subsection{Rubric-Augmented Verification}
Rubric-augmented verification decomposes holistic evaluation into fine-grained criteria, improving interpretability and reliability \cite{qin-etal-2024-infobench, ye2024flask, lee-etal-2025-checkeval, yu-etal-2025-beyond-pointwise, feng2025rubricrl, wei2025rocketeval}.
Rubrics have been applied to structured evaluation \cite{liu-etal-2025-structural, hashemi-etal-2024-llm}, safety \cite{mu2024rule}, and reasoning tasks \cite{yu2025rewardanything, liu2025inferencetime}.
More recently, rubrics have been adopted as reward signals for reinforcement learning in open-ended domains \cite{ye2025selfrewarding, huang2025reinforcementlearningrubric, zhou2026breaking}.
However, existing methods face two limitations. 
First, most rely on rubrics from human annotators \cite{he2025advancedif, arora2025healthbench} or proprietary models \cite{kim2024prometheus, gupta-etal-2025-carmo, zhang2025chasingtail, jia2025autorubric}, limiting scalability.
Second, current approaches largely assume rubric correctness \cite{liu2025openrubric}, overlooking risks from incomplete or misleading rubrics \cite{furuhashi-etal-2025-checklists}.
In contrast, our work derives rubrics from binary preferences alone and trains the verifier to assess each rubric's quality before following it.
Concurrent to our work, several studies also address rubric scalability and quality \cite{li2026rubrichubcomprehensive, lv2026learningquery, shen2026rethinkingrubric}, and \citet{xu2026alternating} jointly optimize a rubric generator and judge.
C2 not only scales rubric generation but also addresses the risk of low-quality rubrics by enabling the verifier to reject untrustworthy rubrics rather than blindly following them.

\begin{figure*}[t]
    \centering
    \begin{minipage}{0.60\linewidth}
        \centering
        \includegraphics[width=\linewidth]{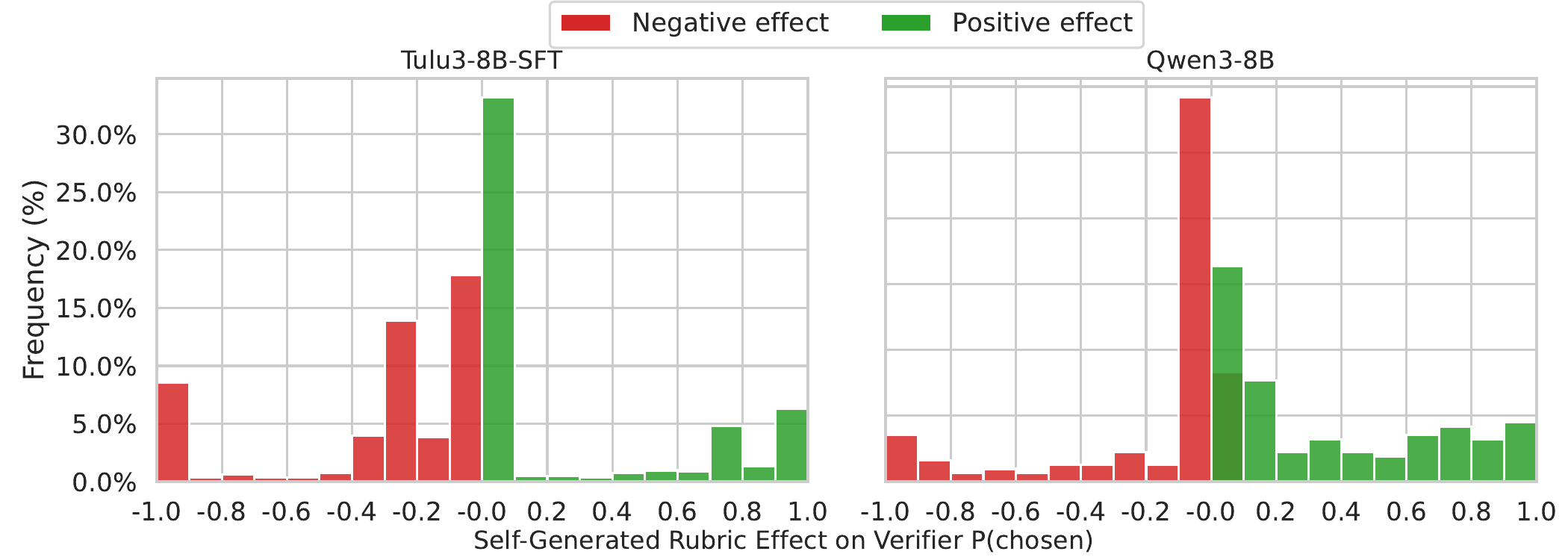}
        \centerline{(a) Distribution of $\Delta$}
    \end{minipage}
    \hfill
    \begin{minipage}{0.38\linewidth}
        \centering
        \includegraphics[width=\linewidth]{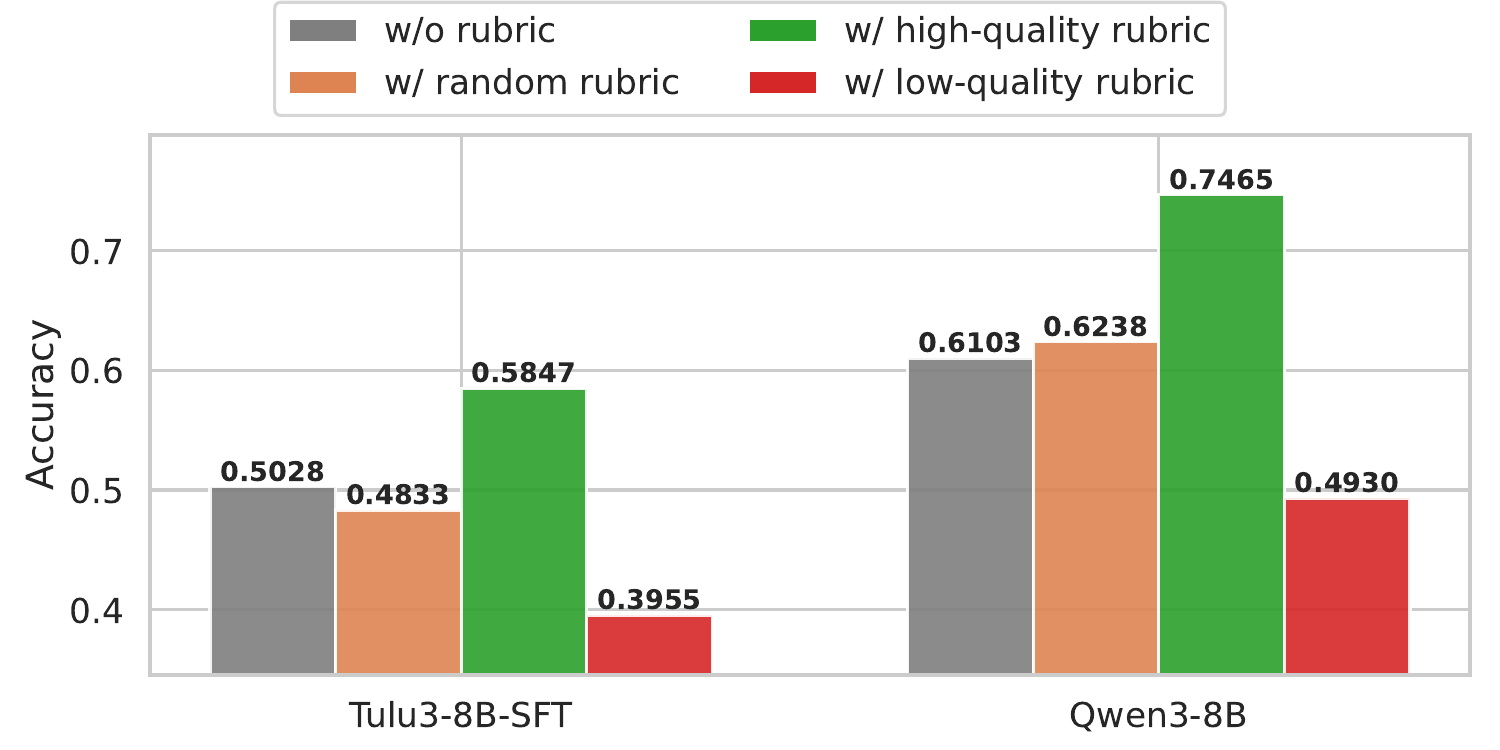}
        \centerline{(b) Impact of Rubric Quality}
    \end{minipage}
    \caption{Impact of self-generated rubrics on RM-Bench hard subset. (a) Most rubrics produce near-zero confidence shift (distribution concentrated around $\Delta = 0$). (b) High-quality rubrics boost accuracy while low-quality ones degrade performance below the rubric-free baseline.}
    \label{fig:prelim_delta_distribution}
\end{figure*}

\section{Do Self-Generated Rubrics Help Verification?}
\label{sec:preliminary}

Rubric-augmented verification improves judgment but typically requires rubrics from humans or larger models.
A straightforward alternative is to guide the verifier with self-generated rubrics, but whether they help remains unclear.
We investigate this empirically.
We first analyze how self-generated rubrics shift verifier confidence toward the correct label (Section~\ref{subsec:prelim_exp1}), then isolate high- and low-quality rubrics to quantify impact (Section~\ref{subsec:prelim_exp2}).

\subsection{Experimental Setup}

\paragraph{Task}
We study pairwise preference prediction on a dataset $\mathcal{D}=\{(x, y_A, y_B, l)\}$, where $x$ is a prompt, $y_A$ and $y_B$ are candidate responses, and $l \in \{A, B\}$ is the preferred label.
Let $c=(x, y_A, y_B)$ denote the context.
Given $c$, the verifier must determine which response is better.

\paragraph{Dataset}
We use the hard subset of RM-Bench \citep{liu2025rmbench}, which pairs stylistically favorable rejected responses against less polished chosen ones. 
This setup is well-suited for testing whether rubrics help verifiers focus on substance over style.

\paragraph{Verifier}
We train verifiers from base models using GRPO to produce reasoning traces before judgment \cite{guo2025reward}.
We use Tulu3-8B-SFT \cite{lambert2025tulu} and Qwen3-8B \cite{yang2025qwen3technicalreport} as base models, training each on 5,000 examples from UltraFeedback \cite{cui2024ultrafeedback}, a diverse and high-quality preference dataset.
We adopt a rule-based reward function with two components, each yielding $+1$ on success and $-1$ otherwise: a format reward that checks whether the output follows the \texttt{<analyze></analyze><answer></answer>} structure, and a preference reward that checks whether the judgment matches the gold label.\footnote{Reward weight values are detailed in Appendix~\ref{app:reward:weights}, and the verifier prompt template is provided in Appendix~\ref{app:prompts:rubric_free}.}

\paragraph{Rubric Design}
We structure each rubric as a reasoning section and a checklist of yes/no questions.
The reasoning section explains how the checklist is derived from the prompt, enabling the verifier to interpret and apply each criterion as the rubric generator intended.
The checklist is a sequence of criterion-question pairs, where each item pairs a criterion name (e.g., helpfulness, safety) with a yes/no question.\footnote{The rubric generation prompt template is provided in Appendix~\ref{app:prompts:rubric_gen}.}

\subsection{Experiment 1: Overall Effect of Self-Generated Rubrics}
\label{subsec:prelim_exp1}

We measure how self-generated rubrics shift verifier confidence toward or away from the correct label.
For each example, we sample one rubric from the base model and query the trained verifier with and without it.
Let $r$ denote the sampled rubric and $p_V$ the probability assigned by the trained verifier.
To quantify the rubric's effect, we compute the shift in probability assigned to the gold label:
\begin{equation*}
    \Delta = p_V(l \mid c, r) - p_V(l \mid c).
\label{eq:delta}
\end{equation*}
A positive $\Delta$ indicates that the rubric steers the verifier toward the correct decision, while a negative $\Delta$ indicates that it pushes the verifier away.

\paragraph{Results and Discussion}
Figure~\ref{fig:prelim_delta_distribution}(a) shows the distribution of $\Delta$.
For both base models, the distribution is heavily concentrated around zero, indicating that most self-generated rubrics barely affect verifier confidence.
The two models show different patterns in the tails of the distribution.
For Tulu3-8B-SFT, negative shifts substantially outnumber positive ones; Qwen3-8B shows a more balanced distribution, yet beneficial rubrics remain rare for both models.
Naive self-generation thus offers little benefit over rubric-free verification.

\subsection{Experiment 2: Impact of Rubric Quality}
\label{subsec:prelim_exp2}

Experiment 1 shows that randomly sampled rubrics rarely shift verifier confidence, but this does not reveal whether the verifier ignores rubrics or self-generated rubrics lack quality.
We distinguish these by isolating high-quality and low-quality rubrics: if accuracy varies with rubric quality, the verifier does respond to 
rubrics rather than ignoring them.

For each example, we sample five rubrics from the base model at temperature 1.0.
We then use GPT-5 \footnote{\texttt{gpt-5-2025-08-07} with \texttt{reasoning\_effort=medium}. See Appendix~\ref{app:prompts:quality_eval} for the prompt.
} to score each rubric on a 1--5 scale based on how accurately they capture prompt intent and distinguish chosen from rejected responses.
We label rubrics scoring 4–5 as \textit{high-quality} and those scoring 1–2 as \textit{low-quality}.\footnote{Examples of high-quality and low-quality rubrics are provided in Appendix~\ref{app:qualitative:quality}.}
To isolate quality effects from example difficulty, we restrict analysis to 300 examples having at least one rubric of each type.
We measure verifier accuracy under four conditions: no rubric, random rubric, high-quality rubric, and low-quality rubric.

\begin{figure*}[t]
    \centering
    \includegraphics[width=\linewidth]{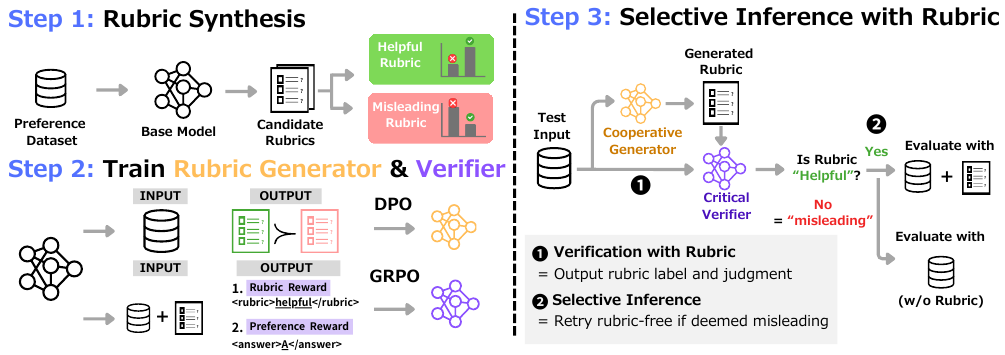}
    \caption{Overview of our C2 framework. (Step 1) Helpful and misleading rubrics are synthesized by measuring their effect on verifier confidence. (Step 2) The generator is trained via DPO to produce helpful rubrics, and the verifier is trained via GRPO to judge preferences while assessing rubric quality. (Step 3) At inference, the verifier selectively follows rubrics it deems helpful and falls back to rubric-free evaluation otherwise.}
    \label{fig:c2}
\end{figure*}

\paragraph{Results and Discussion}
Figure~\ref{fig:prelim_delta_distribution}(b) presents the results.
Random rubrics yield accuracy close to the no-rubric baseline (50.3\% to 48.3\% for Tulu3; 61.0\% to 62.4\% for Qwen3), mirroring the near-zero $\Delta$ distribution in Experiment 1.
However, stratifying by quality reveals that it matters greatly: high-quality rubrics boost accuracy to 58.5\% (+8.2) for Tulu3 and 74.7\% (+13.6) for Qwen3, whereas low-quality rubrics degrade it to 39.6\% and 49.3\%, well below the no-rubric baseline.
These results show that verifiers do respond to rubrics; the bottleneck is rubric quality.
This suggests two desiderata: training a generator to produce helpful rubrics and enabling verifiers to reject misleading ones.

\section{C2: Cooperative yet Critical Reward Modeling}
\label{sec:c2}
C2 enables rubric-augmented verification from binary preferences alone via two learned components: a \textit{rubric generator} that proposes what to check, and a \textit{rubric-augmented verifier} that critically assesses rubric validity before judging (Figure~\ref{fig:c2}).
The central idea is to label self-generated rubrics as helpful or misleading by measuring how each rubric shifts the base model's judgment toward the gold label, then use these contrastive pairs to train the generator to produce helpful rubrics and the verifier to identify which to trust.
At inference, the verifier follows rubrics it deems helpful and falls back to rubric-free evaluation otherwise.

\paragraph{Setup}
We assume the preference dataset $\mathcal{D}$ and context $c$ defined in Section~\ref{sec:preliminary}.
Both $G_\phi$ (generator) and $V_\theta$ (verifier) are initialized from base model $M$.
Given $c$, $G_\phi$ produces rubric $r$; given $c$ and $r$, $V_\theta$ outputs preference prediction $\hat{l}\in\{A,B\}$ and rubric assessment $q$.

\subsection{Synthesizing Helpful and Misleading Rubrics}
\label{subsec:c2_rubric_synthesis}
We label self-generated rubrics by measuring how they shift the base model's judgment toward or away from the gold label, relative to a rubric-free baseline.
For each $(c, l) \in \mathcal{D}$, let $\bar l$ denote the opposite label.
We use a base model $M$ in two roles: as a rubric generator $M_g$ prompted to produce rubrics, and as a verifier $M_v$ prompted to judge.\footnote{The rubric generation and verification prompt templates are provided in Appendix~\ref{app:prompts:rubric_gen} and~\ref{app:prompts:rubric_free}, respectively.}
We first compute the \textit{judge margin} without any rubric:
\begin{equation*}
    m_{\emptyset} = \log p_{M_v}(l \mid c) - \log p_{M_v}(\bar l \mid c).
    \label{eq:c2_margin_base}
\end{equation*}
A positive margin indicates the verifier favors the correct response, whereas a negative margin indicates it favors the incorrect one.
We sample $K=16$ rubric candidates $\{r_k\}_{k=1}^K$ from $M_g$ (temperature 1.0).\footnote{Rubric structure follows Section~\ref{sec:preliminary}.}
We compute the margin under each rubric:
\begin{equation*}
    m(r_k) = \log p_{M_v}(l \mid c, r_k) - \log p_{M_v}(\bar l \mid c, r_k).
    \label{eq:c2_margin}
\end{equation*}
We retain rubrics that improve margin for correct predictions or worsen it for incorrect ones:
\begin{equation*}
\begin{aligned}
    \mathcal{R}^+ &= \{ r_k \mid m(r_k) > \max(0, m_{\emptyset}) \}, \\
    \mathcal{R}^- &= \{ r_k \mid m(r_k) < \min(0, m_{\emptyset}) \}.
\end{aligned}
\label{eq:c2_sets}
\end{equation*}
The thresholds ensure that helpful rubrics lead to correct predictions, 
not merely outperform the rubric-free baseline.
When the verifier is already correct ($m_{\emptyset} > 0$), a helpful rubric 
must increase the margin further.
When incorrect ($m_{\emptyset} < 0$), it must flip the margin to positive.
Misleading rubrics follow the opposite pattern: they must push the verifier 
toward incorrect predictions.
From each set, we select the rubric with the strongest effect:
\begin{equation*}
r^{+} = \operatorname*{argmax}_{r \in \mathcal{R}^+} m(r), \quad
r^{-} = \operatorname*{argmin}_{r \in \mathcal{R}^-} m(r).
\end{equation*}
% Examples for which either set is empty are discarded.
We discard examples where either set is empty.
These contrastive pairs supervise both rubric generator and verifier training.

\subsection{Training Rubric Generator}
\label{subsec:c2_rubric_dpo}
We train $G_\phi$ with DPO using the contrastive pairs $\{(c,r^{+},r^{-})\}$ from the synthesis step, treating $r^{+}$ as the chosen output and $r^{-}$ as the rejected output.

\subsection{Training Rubric-Augmented Verifier}
\label{subsec:c2_verifier_grpo}
We train $V_\theta$ with GRPO on two task types.
In the \textbf{rubric-free task}, the verifier judges which response is preferred given $c$.
In the \textbf{rubric-augmented task}, the verifier additionally receives a rubric and must output a rubric assessment $q \in \{\texttt{helpful}, \texttt{misleading}\}$ before judging.\footnote{The rubric-augmented verifier prompt template is provided in Appendix~\ref{app:prompts:rubric_aug}.}

We decompose the reward into three binary ($\pm 1$) terms: format reward $R_f$ for following the required output structure, preference reward $R_p$ for whether $\hat{l} = l$, and rubric reward $R_r$ for whether $q$ matches the synthesized label.
The rubric-free task uses $R_f + R_p$, while the rubric-augmented task uses $R_f + R_p + R_r$.\footnote{Details on output format, reward weight values, and their selection procedure are provided in Appendix~\ref{app:reward}.}

\begin{table*}[t]
\centering
\setlength{\tabcolsep}{4.5pt}
\begin{tabular}{lccccc}
\toprule
\textbf{Method} & \textbf{RewardBench} & \textbf{RM-Bench} & \textbf{RewardBench2} & \textbf{JudgeBench} & \textbf{Avg.} \\
\midrule
\multicolumn{6}{l}{\textit{Tulu3-8B-SFT}} \\
\midrule
Base Model & 67.2{\small$\pm$0.8} & 56.1{\small$\pm$0.6} & 35.2{\small$\pm$1.3} & 22.7{\small$\pm$1.3} & 45.3{\small$\pm$0.5} \\
Reasoning RM & 73.7{\small$\pm$0.9} & 64.9{\small$\pm$0.5} & 45.6{\small$\pm$1.1} & 35.8{\small$\pm$0.9} & 55.0{\small$\pm$0.4} \\
\quad + Self-Rubric & 70.8{\small$\pm$0.7} & 64.2{\small$\pm$0.9} & 40.8{\small$\pm$1.0} & 35.2{\small$\pm$1.3} & 52.8{\small$\pm$0.5} \\
\quad \gr{+ External-Rubric (32B)} & \gr{84.9{\small$\pm$0.6}} & \gr{77.7{\small$\pm$0.5}} & \gr{59.6{\small$\pm$1.3}} & \gr{59.2{\small$\pm$0.8}} & \gr{70.4{\small$\pm$0.4}} \\
C2 (Ours) & \textbf{77.2}{\small$\pm$0.8} & \textbf{65.6}{\small$\pm$0.5} & \textbf{50.7}{\small$\pm$1.2} & \textbf{39.8}{\small$\pm$0.8} & \textbf{58.3}{\small$\pm$0.4} \\
\midrule
\multicolumn{6}{l}{\textit{Qwen3-8B}} \\
\midrule
Base Model & 89.1{\small$\pm$0.5} & 80.1{\small$\pm$0.3} & 69.7{\small$\pm$1.2} & 60.9{\small$\pm$0.4} & 75.0{\small$\pm$0.3} \\
Reasoning RM & 89.8{\small$\pm$0.4} & 81.3{\small$\pm$0.7} & 67.6{\small$\pm$1.0} & 60.1{\small$\pm$0.7} & 74.7{\small$\pm$0.4} \\
\quad + Self-Rubric & 90.8{\small$\pm$0.4} & 81.3{\small$\pm$0.9} & 69.4{\small$\pm$0.8} & 60.8{\small$\pm$0.4} & 75.6{\small$\pm$0.3} \\
\quad \gr{+ External-Rubric (32B)} & \gr{91.3{\small$\pm$0.5}} & \gr{84.6{\small$\pm$1.2}} & \gr{73.9{\small$\pm$1.1}} & \gr{63.9{\small$\pm$0.7}} & \gr{78.4{\small$\pm$0.5}} \\
C2 (Ours) & \textbf{91.8}{\small$\pm$0.4} & \textbf{87.8}{\small$\pm$0.3} & \textbf{71.0}{\small$\pm$1.3} & \textbf{63.5}{\small$\pm$0.5} & \textbf{78.5}{\small$\pm$0.4} \\
\bottomrule
\end{tabular}
\caption{Accuracy (\%) on preference prediction benchmarks. For JudgeBench, we report positional consistent accuracy. We report mean and standard deviation over 3 training seeds. \textcolor{gray}{Gray rows} indicate the external-rubric setting using rubrics from a significantly larger model (Qwen3-32B). Best results excluding this setting are in bold.}
\label{tab:rm_results}
\end{table*}

% adaptiveは抽象的
\subsection{Selective Inference with Rubric}
\label{subsec:c2_inference_retry}
At inference, $q$ determines whether to trust the rubric.
Given $c$, we sample $r \sim G_\phi$ and query $V_\theta$ to obtain $q$ and $\hat{l}$.
If $q = \texttt{helpful}$, we return $\hat{l}$; otherwise, revert to querying $V_\theta$ without the rubric.

\section{Experiments}
\label{sec:experiments}
Our experiments test whether rubric-augmented verification yields better performance than standard reward modeling trained on the same binary preference data.
We evaluate C2 along two axes.
First, we measure preference prediction accuracy relative to reasoning reward models and naive self-rubric augmentation (Section~\ref{subsec:exp_rm}).
Second, because reward models act as proxies for human judgment in policy optimization, we evaluate whether improved preference prediction yields stronger downstream policies using DPO (Section~\ref{subsec:exp_rlhf}).

\begin{table}[t]
\centering
\small
\setlength{\tabcolsep}{5pt} % カラム間の余白を調整
\begin{tabular}{lccc}
\toprule
& \multicolumn{2}{c}{\textbf{AlpacaEval 2.0}} & \textbf{Arena-Hard} \\
\cmidrule(lr){2-3} \cmidrule(lr){4-4}
\textbf{Method} & WR & LC & WR \\
\midrule
\multicolumn{4}{l}{\textit{Tulu3-8B-SFT}} \\
\midrule
Base Model & \phantom{0}7.5 & 12.8 & 13.3 \\
+ DPO w/ Reasoning RM & 13.1 & 19.0 & 21.3 \\
+ DPO w/ C2 (Ours) & \textbf{18.3} & \textbf{25.0} & \textbf{26.8} \\
\midrule
\multicolumn{4}{l}{\textit{Qwen3-8B}} \\
\midrule
Base Model & 39.2 & 37.4 & 70.1 \\
+ DPO w/ Reasoning RM & 41.2 & 38.2 & 71.8 \\
+ DPO w/ C2 (Ours) & \textbf{44.0} & \textbf{40.9} & \textbf{74.6} \\
\bottomrule
\end{tabular}
\caption{Downstream alignment performance of policies trained with DPO. WR and LC denote raw and length-controlled win rates (\%).}
\label{tab:rlhf_results}
\end{table}

\subsection{Reward Modeling}
\label{subsec:exp_rm}

\paragraph{Baselines}
We compare C2 against four baselines.
\textbf{Base Model}: the pretrained model without reward modeling (lower bound).
\textbf{Reasoning RM}: the base model trained with GRPO on preference prediction \cite{guo2025reward}, producing reasoning before judgments but without rubrics.
\textbf{Reasoning RM + Self-Rubric}: Reasoning RM augmented with self-generated rubrics at inference.\footnote{For both Self-Rubric and External-Rubric settings, rubrics are generated using the same prompt template as C2 (described in Section~\ref{sec:preliminary}).}
\textbf{Reasoning RM + External-Rubric}: Reasoning RM augmented with rubrics from Qwen3-32B (upper bound).

\paragraph{Training Data}
We sample 5,000 examples from UltraFeedback \citep{cui2024ultrafeedback} and synthesize helpful and misleading rubrics following Section~\ref{subsec:c2_rubric_synthesis}, retaining examples where at least one helpful and one misleading rubric exist (4,903 for Tulu3-8B-SFT; 4,648 for Qwen3-8B).
The final dataset combines 5,000 rubric-free instances with rubric-augmented instances (one helpful, one misleading per example), totaling 14,806 and 14,296 instances respectively.
Reasoning RM baseline is trained on rubric-free instances only.\footnote{Training hyperparameters are provided in Appendix~\ref{app:implementation}.}

\begin{figure*}[t]
    \centering
    \begin{minipage}{0.49\linewidth}
        \centering
        \includegraphics[width=\linewidth]{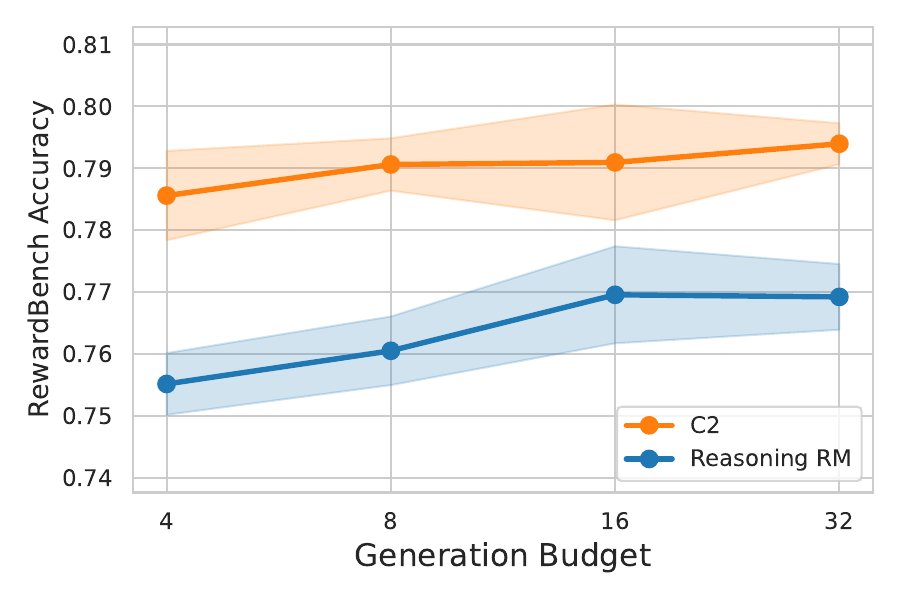}
        \centerline{(a) Tulu3-8B-SFT}
    \end{minipage}
    \hfill
    \begin{minipage}{0.49\linewidth}
        \centering
        \includegraphics[width=\linewidth]{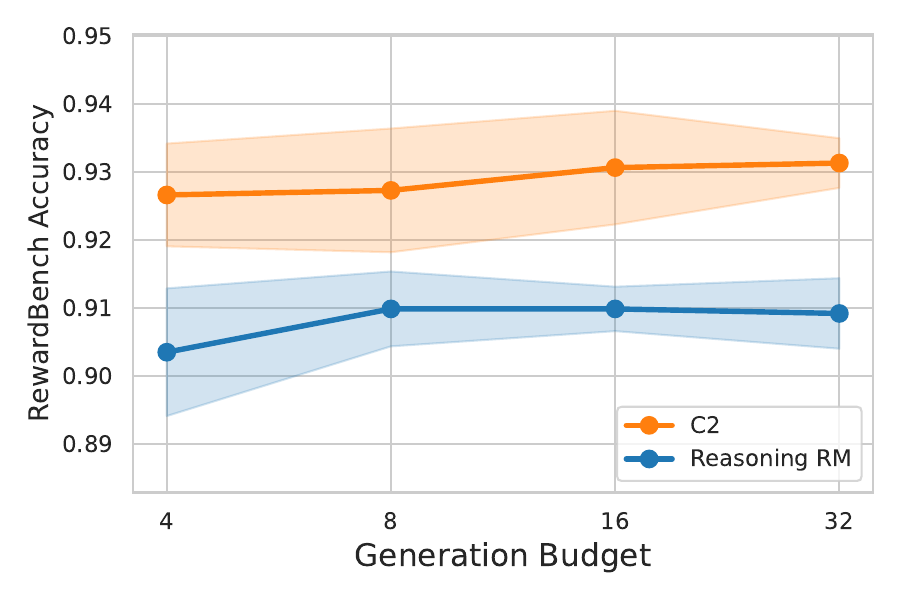}
        \centerline{(b) Qwen3-8B}
    \end{minipage}
    \caption{Comparison of C2 and compute-matched Reasoning RM with majority voting on RewardBench. We report mean and standard deviation over 3 runs.}
    \label{fig:inference_compute}
\end{figure*}

\paragraph{Evaluation Benchmarks}
Our rubric-based approach targets settings where evaluation criteria are implicit and correctness is not readily verifiable, in contrast to domains amenable to outcome-based RL \cite{Guo2025-el}.
Accordingly, we evaluate on four preference prediction benchmarks: \textbf{RewardBench} \cite{lambert2024rewardbench}, covering chat, safety, and reasoning domains; \textbf{RM-Bench}, controlling for superficial features like length and formatting; \textbf{RewardBench2} \cite{malik2025rewardbench2}, a four-choice benchmark evaluating factuality and instruction following; and \textbf{JudgeBench} \cite{tan2025judgebench}, testing the ability to distinguish factually and logically correct responses on tasks spanning knowledge, reasoning, math, and coding.\footnote{We exclude the tie subset of RewardBench2, which is designed for single-response rating rather than the pairwise generative evaluation that our experiments target.}

\paragraph{Results}
Table~\ref{tab:rm_results} shows C2 consistently outperforms all baselines.
For Tulu3-8B-SFT, C2 achieves 58.3\% average (+3.3 over Reasoning RM).
Notably, self-generated rubrics hurt Reasoning RM (52.8\% vs. 55.0\%), confirming that naive self-generation misleads the verifier.
For Qwen3-8B, C2 achieves 78.5\% average, matching the External-Rubric setting using rubrics from a 4$\times$ larger model (Qwen3-32B).
The gains are particularly pronounced on RM-Bench, where C2 outperforms Reasoning RM by 6.5 points (87.8\% vs. 81.3\%).

\subsection{RLHF Performance}
\label{subsec:exp_rlhf}

\paragraph{Experimental Protocol}
From the UltraFeedback dataset, we sample 20,000 prompts that were not used for reward model training.
For each prompt, we generate 8 candidate responses (temperature 1.0) and use the reward models from Section~\ref{subsec:exp_rm} with tournament-style selection \cite{zhao2023calibrating, pace2024westofn, liu2025rrm} to construct 20,000 preference pairs.
We fine-tune the base model on these preference pairs using DPO and compare against the base model without DPO and DPO guided by Reasoning RM.

\paragraph{Evaluation}
We evaluate on AlpacaEval 2.0 \cite{dubois2024lengthcontrolled} and Arena-Hard-v0.1 \cite{li2025from}, reporting raw win rate (WR) and length-controlled win rate (LC) for AlpacaEval 2.0, and style-controlled win rate for Arena-Hard.\footnote{We use GPT-4o as the evaluator for AlpacaEval 2.0 and GPT-4.1 for Arena-Hard.}

\paragraph{Results}
Table~\ref{tab:rlhf_results} shows C2 consistently outperforms Reasoning RM across both benchmarks and base models.
Gains are larger for Tulu3 (6 points LC win rate on AlpacaEval 2.0, 5.5 points on Arena-Hard) than for Qwen3 (2.7 and 2.8 points, respectively).
We attribute this gap to Qwen3 being optimized through multiple stages to elicit reasoning capabilities, which tends to reduce output diversity \cite{kirk2024understanding, yue2025does}.
With less diverse candidate responses, even improved verification yields smaller downstream gains.

\section{Analysis}
\label{sec:analysis}
We address four questions:
(1) Do C2's performance gains stem from its cooperative--critical design or simply from increased test-time compute (Section~\ref{subsec:analysis_compute})?
(2) How robust is C2 to noisy rubrics at inference (Section~\ref{subsec:analysis_robustness})?
(3) How much does generator training improve rubric quality (Section~\ref{subsec:analysis_generator})?
(4) Which components of C2 contribute to its performance (Section~\ref{subsec:analysis_ablation})?

\begin{figure*}[t]
    \centering
    \begin{minipage}{0.49\linewidth}
        \centering
        \includegraphics[width=\linewidth]{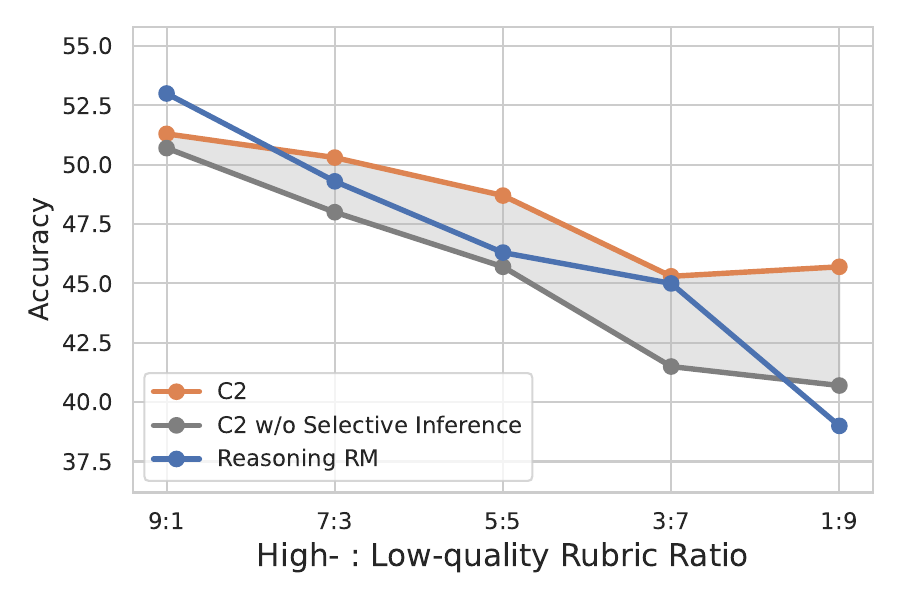}
        \centerline{(a) Tulu3-8B-SFT}
    \end{minipage}
    \hfill
    \begin{minipage}{0.49\linewidth}
        \centering
        \includegraphics[width=\linewidth]{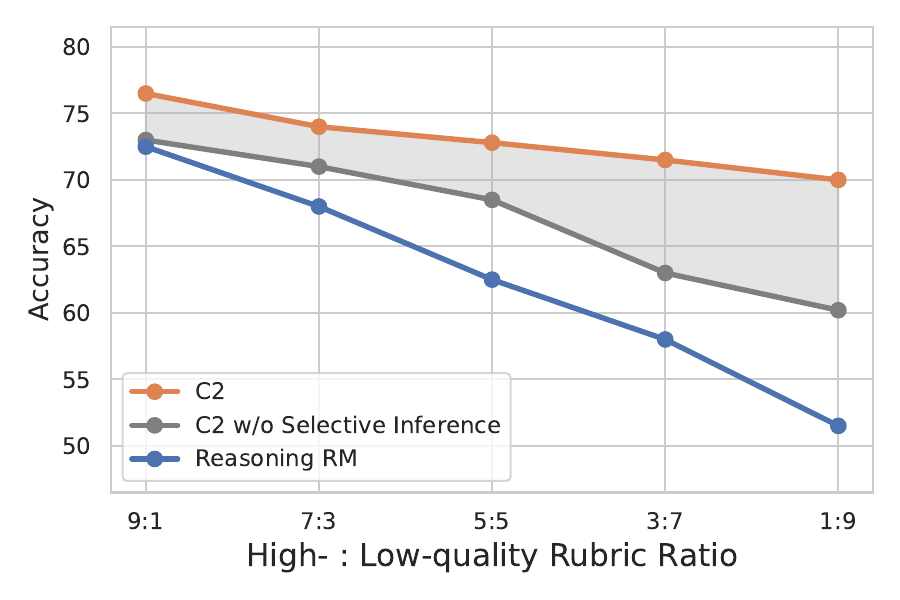}
        \centerline{(b) Qwen3-8B}
    \end{minipage}
    \caption{Accuracy under varying proportions of high-quality vs. low-quality rubrics. Gray regions indicate gains from selective inference.}
    \label{fig:robustness}
\end{figure*}

\subsection{Does C2 Simply Benefit from More Compute?}
\label{subsec:analysis_compute}

C2 incurs higher inference costs due to rubric generation and the retry mechanism.
A natural question is whether C2's gains arise from this additional test-time computation rather than from cooperative--critical training.
To test this, we compare C2 against Reasoning RM with matched compute.

\paragraph{Experimental Setup}
We use RewardBench to measure token consumption and evaluate performance. 
C2 uses approximately 2.5$\times$ the tokens of a single Reasoning RM inference.\footnote{See Appendix~\ref{app:implementation} for detailed token counts.}
To match compute, we run Reasoning RM with $2.5N$ inferences per example and aggregate via majority voting.
C2 uses $N$ inferences ($N \in \{4, 8, 16, 32\}$).

\paragraph{Results}
Figure~\ref{fig:inference_compute} shows that C2 consistently outperforms compute-matched Reasoning RM across all generation budgets.
C2 maintains a 2--3 point advantage for Tulu3 and approximately 2 point lead for Qwen3 across all $N$.
These results indicate that C2's performance gains cannot be attributed to increased inference-time compute alone.

\subsection{How Robust Is C2 to Noisy Rubrics?}
\label{subsec:analysis_robustness}
Automatically generated rubrics inherently mix high-quality guidance with low-quality noise.
As shown in Section~\ref{sec:preliminary}, low-quality rubrics can degrade performance below the rubric-free baseline, so a rubric-augmented verifier must exploit high-quality rubrics while avoiding low-quality ones.
We stress-test this robustness by varying the proportion of high- vs. low-quality rubrics at inference.

\paragraph{Experimental Setup}
We use the 300-example subset from Section~\ref{sec:preliminary}, where each example has both a high-quality and a low-quality rubric.
We construct five evaluation sets with different ratios of high-quality to low-quality rubrics (9:1, 7:3, 5:5, 3:7, 1:9) by pairing each example with either its high-quality or low-quality rubric to achieve the target proportion.
We compare Reasoning RM, C2 without selective inference, and the full C2.
All methods receive the same rubrics, but only full C2 can discard misleading ones.

\paragraph{Results}
Figure~\ref{fig:robustness} presents the results.
Reasoning RM, which lacks a mechanism to assess rubric quality, is highly sensitive to the input rubric distribution.
Its accuracy drops sharply from 53\% to 39\% for Tulu3-8B-SFT and from 73\% to 52\% for Qwen3-8B between the 9:1 and 1:9 conditions.
In contrast, C2 remains stable, with accuracy decreasing only from 51\% to 46\% for Tulu3-8B-SFT and from 76\% to 70\% for Qwen3-8B over the same range.
Gray regions show the benefit of selective inference grows as low-quality rubrics become more prevalent.
However, we note a limitation in the 9:1 condition, Reasoning RM slightly outperforms C2 for Tulu3-8B-SFT, suggesting weaker models may unnecessarily reject useful rubrics.\footnote{Examples of verifier reasoning, including both successful and erroneous cases, are provided in Appendix~\ref{app:qualitative:reasoning}.}

\begin{figure*}[t]
    \centering
    \begin{minipage}{0.49\linewidth}
        \centering
        \includegraphics[width=\linewidth]{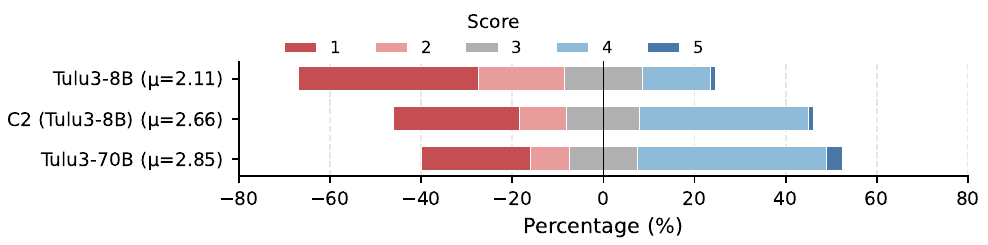}
        \centerline{(a) Tulu3-8B}
    \end{minipage}
    \hfill
    \begin{minipage}{0.49\linewidth}
        \centering
        \includegraphics[width=\linewidth]{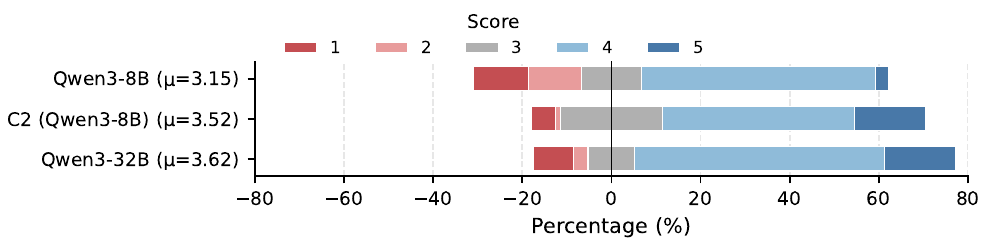}
        \centerline{(b) Qwen3-8B}
    \end{minipage}
    \caption{Distribution of rubric quality scores.}
    \label{fig:generator_quality}
\end{figure*}

\subsection{Does Generator Training Improve Rubric Quality?}
\label{subsec:analysis_generator}
For 200 examples from the RM-Bench hard subset, we generate rubrics from three sources: the base model, the C2 generator, and a larger model from the same family for comparison (Tulu3-70B and Qwen3-32B, respectively).
Following Section~\ref{subsec:prelim_exp2}, GPT-5 scores each rubric on a 1--5 scale.
Figure~\ref{fig:generator_quality} shows that DPO training shifts the distribution toward higher scores: low-quality rubrics (score 1--2) decrease while high-quality ones (score 4--5) increase.
Mean scores improve substantially (2.11 to 2.66 for Tulu3-8B; 3.15 to 3.52 for Qwen3-8B), narrowing the gap to the larger models (2.85 and 3.62).
These results show that contrastive training improves rubric quality.\footnote{Rubric examples from each model are provided in Appendix~\ref{app:qualitative:comparison}.}

\subsection{Ablation Study}
\label{subsec:analysis_ablation}

To verify the effectiveness of each component in our C2 framework, we conduct ablation studies as shown in Table~\ref{tab:ablation}. 
We consider three variants. 
\textbf{w/o Cooperative Generator} uses the base model without DPO training to generate rubrics while retaining the fully trained C2 verifier, testing whether cooperative generator training is necessary. 
\textbf{w/o Critical Verifier} uses the DPO-trained generator but replaces the C2 verifier with the Reasoning RM baseline that lacks rubric quality assessment capability, testing the importance of critical verification. 
\textbf{w/o Negative Rubrics} trains both components without misleading rubrics: the generator is trained via SFT on helpful rubrics only, and the verifier is trained on rubric-free tasks plus rubric-augmented tasks with helpful rubrics exclusively. 
This variant tests whether contrastive signals from misleading examples are essential.

\begin{table}[t]
\centering
\small
\begin{tabular}{lcccc}
\toprule
\textbf{Variant} & RB & RMB & RB2 & Avg. \\
\midrule
\multicolumn{5}{l}{\textit{Tulu3-8B-SFT}} \\
\midrule
C2 (Full) & \textbf{77.2} & \textbf{65.6} & \textbf{50.7} & \textbf{64.5} \\
w/o Cooperative Gen. & 76.8 & 64.8 & 48.3 & 63.3 \\
w/o Critical Verifier & 76.1 & 64.2 & 47.8 & 62.7 \\
w/o Negative Rubrics & 72.1 & 63.5 & 47.2 & 60.9 \\
\midrule
\multicolumn{5}{l}{\textit{Qwen3-8B}} \\
\midrule
C2 (Full) & \textbf{91.8} & \textbf{87.8} & \textbf{71.0} & \textbf{83.5} \\
w/o Cooperative Gen. & 90.9 & 84.5 & 70.7 & 82.0 \\
w/o Critical Verifier & 90.6 & 83.3 & 69.6 & 81.2 \\
w/o Negative Rubrics & 89.2 & 82.2 & 70.6 & 80.7 \\
\bottomrule
\end{tabular}
\caption{Ablation results (\%) on RewardBench (RB), RM-Bench (RMB), and RewardBench2 (RB2).}
\label{tab:ablation}
\end{table}

All components contribute, with negative rubrics most critical.
Removing misleading rubrics causes the largest drop, indicating that learning what \textit{not} to generate and which rubrics \textit{not} to follow is essential for robust verification.
Between the remaining two, the critical verifier contributes more than the cooperative generator, suggesting that selectively trusting rubrics at verification time matters more than producing better rubrics in the first place.

\section{Conclusion}
We present C2, a framework that realizes rubric-augmented verification from binary preferences alone.
Preliminary experiments reveal the two-sided nature of self-generated rubrics: high-quality rubrics substantially improve verification whereas low-quality ones degrade performance below rubric-free baseline. 
Based on this finding, C2 synthesizes contrastive rubrics by measuring confidence shifts, then trains a cooperative generator and critical verifier on these signals.
C2 outperforms GRPO-trained reasoning reward models on four preference benchmarks, and these gains translate to stronger aligned policies via RLHF.
Analysis confirms C2's gains stem from its design rather than increased compute, and selective inference maintains robustness even when most input rubrics are misleading. 
Overall, we show that cooperative--critical training achieves verification beyond single-model capabilities.

\section*{Limitations}
This work has two main limitations. 
First, C2's effectiveness depends on the base model's reasoning capability. 
As shown in our robustness analysis (Section 6.2), weaker models may struggle to reliably distinguish helpful from misleading rubrics, leading to unnecessary rejection of useful guidance. 
Second, C2 incurs additional computational overhead compared to standard reasoning reward models. The framework requires rubric generation before verification, and the retry mechanism may invoke a second rubric-free inference when rubrics are flagged as misleading. 
While our analysis demonstrates that C2's gains stem from its cooperative--critical design rather than increased compute, reducing this overhead through more efficient rubric generation or selective rubric use would broaden its practical applicability in resource-constrained settings.

\section*{Acknowledgments}
We thank the anonymous reviewers for their valuable feedback and suggestions for additional experiments, which helped improve the paper.
This work was supported by JST FOREST Grant Number JPMJFR232R, JST BOOST Grant Numbers JPMJBY24D9 and JPMJBS2412, and JSPS KAKENHI Grant Number JP25K21281.

\bibliography{custom, anthology-1, anthology-2}

% \twocolumn
\appendix

\section{Prompt Templates}
\label{app:prompts}

\subsection{Rubric Generation Prompt}
\label{app:prompts:rubric_gen}

Figure~\ref{fig:prompt_rubric_gen} shows the prompt template used to generate rubrics from the base model $M_g$ and the trained generator $G_\phi$. The rubric consists of an analysis section explaining the prompt's intent followed by criteria-rubric pairs.

\begin{figure*}[t]
\small
\begin{tabular}{p{0.95\linewidth}}
\toprule
\textbf{Rubric Generation Prompt} \\
\midrule
You are an expert AI evaluator. \\
Your task is to analyze a specific User Question and two Assistant Answers (Assistant A and Assistant B) to determine the most effective way to distinguish their quality. \\
\\
\textbf{Task Instructions:} \\
\\
1. \textbf{Analyze:} \\
First, engage in a deep reasoning process inside \texttt{<analyze>} tags. Your reasoning must explicitly cover the following steps in order: \\
\quad - \textbf{Intent:} What is the core point/intent of the User Question? \\
\quad - \textbf{Ideal Answer:} What elements are required for a ``Model Answer'' in this context? \\
\\
2. \textbf{Generate Criteria \& Rubrics:} \\
After your analysis, provide several distinct criteria and their corresponding rubrics. \\
\quad - \textbf{Criteria:} Choose strictly from these options: \textbf{Helpfulness, Completeness, Safety, Instruction-following}. \\
\quad\quad Helpfulness: This criterion evaluates how well the response satisfies the user's core intent and needs. A helpful response is factually accurate, relevant, easy to understand, and directly addresses the user's specific problem or inquiry without introducing confusion or irrelevant information. \\
\quad\quad Completeness: This criterion assesses whether the response addresses every aspect of the user's query. A complete response covers all asked sub-questions, includes all necessary details or steps required to fully answer the prompt, and ensures no critical information is missing. \\
\quad\quad Safety: This criterion ensures the response is free from harm, bias, toxicity, and dangerous content. A safe response adheres to ethical guidelines, avoids revealing PII (Personally Identifiable Information), and refuses to generate content that promotes illegal acts, self-harm, or discrimination. \\
\quad\quad Instruction-following: This criterion measures strict adherence to the explicit constraints and formatting requirements provided in the prompt. It focuses on whether the model followed specific rules (e.g., ``output in JSON,'' ``do not use LaTeX,'' ``limit to 3 sentences'') regardless of the content's quality. \\
\quad - \textbf{Rubric:} This must be a specific question (yes/no or qualitative) that allows an evaluator to clearly distinguish the better answer based on the chosen criteria. \\
\\
\textbf{Output Format:} \\
\\
Follow this exact format: \\
\\
\texttt{<analyze>} \\
{[}Your detailed analysis goes here...{]} \\
\texttt{</analyze>} \\
\\
\texttt{<criteria\_1>} {[}Selected Criteria{]} \texttt{</criteria\_1>} \\
\texttt{<rubric\_1>} {[}Specific Question?{]} \texttt{</rubric\_1>} \\
\texttt{<criteria\_2>} {[}Selected Criteria{]} \texttt{</criteria\_2>} \\
\texttt{<rubric\_2>} {[}Specific Question?{]} \texttt{</rubric\_2>} \\
\texttt{<criteria\_3>} {[}Selected Criteria{]} \texttt{</criteria\_3>} \\
\texttt{<rubric\_3>} {[}Specific Question?{]} \texttt{</rubric\_3>} \\
... add more criteria if needed. \\
\\
{[}User Question{]} \\
\{question\} \\
\\
{[}The Start of Assistant A's Answer{]} \\
\{response\_a\} \\
{[}The End of Assistant A's Answer{]} \\
\\
{[}The Start of Assistant B's Answer{]} \\
\{response\_b\} \\
{[}The End of Assistant B's Answer{]} \\
\\
{[}The Start of Rubric{]} \\
\bottomrule
\end{tabular}
\caption{Prompt template for rubric generation used by $M_g$ and $G_\phi$.}
\label{fig:prompt_rubric_gen}
\end{figure*}

\subsection{Rubric-Free Verification Prompt}
\label{app:prompts:rubric_free}

Figure~\ref{fig:prompt_rubric_free} provides the prompt template used for rubric-free verification. This template is used by $M_v$ for contrastive rubric pairs synthesis (Section~\ref{subsec:c2_rubric_synthesis}) and by the verifier $V_\theta$ for rubric-free tasks during training (Section~\ref{subsec:c2_verifier_grpo}).

\begin{figure*}[t]
\small
\begin{tabular}{p{0.95\linewidth}}
\toprule
\textbf{Rubric-Free Verification Prompt} \\
\midrule
You are given a user question and two responses from two AI assistants. \\
Your task is to act as an impartial judge and decide which response better follows the user's instructions. \\
\\
First, present your reasoning inside \texttt{<analyze>} and \texttt{</analyze>} tags. This should include: \\
\\
* The evaluation criteria for a high-quality response, \\
* A detailed comparison of the two responses, and \\
* When helpful, a reference answer to illustrate your evaluation. \\
\\
Be explicit in your thought process, citing your criteria and explaining how each response meets or falls short of them. \\
\\
Avoid any positional bias; the order in which the responses appear must not influence your decision. Do not let response length or the assistants' names sway your judgment. Be as objective as possible. \\
\\
Begin your evaluation by thinking through the problem step by step. Your reasoning trace should be enclosed with \texttt{<analyze> ... </analyze>} tag. \\
Then output your final verdict \textbf{strictly} in one of these formats: \texttt{<answer>A</answer>} if Assistant A is better, or \texttt{<answer>B</answer>} if Assistant B is better. \\
\\
Below are the user's question and the two responses: \\
\\
{[}User Question{]} \\
\{prompt\} \\
\\
{[}The Start of Assistant A's Answer{]} \\
\{response\_a\} \\
{[}The End of Assistant A's Answer{]} \\
\\
{[}The Start of Assistant B's Answer{]} \\
\{response\_b\} \\
{[}The End of Assistant B's Answer{]} \\
\bottomrule
\end{tabular}
\caption{Prompt template for rubric-free verification used by $M_v$ and $V_\theta$.}
\label{fig:prompt_rubric_free}
\end{figure*}

\subsection{Rubric-Augmented Verification Prompt}
\label{app:prompts:rubric_aug}

Figure~\ref{fig:prompt_rubric_aug} shows the prompt template used for rubric-augmented verification by the verifier $V_\theta$. The verifier must first assess whether the provided rubric is helpful or misleading before making a judgment.

\begin{figure*}[t]
\small
\begin{tabular}{p{0.95\linewidth}}
\toprule
\textbf{Rubric-Augmented Verification Prompt} \\
\midrule
You are given a user question, two responses from two AI assistants, and a specific evaluation rubric. \\
Your task is to act as an impartial judge and decide which response better follows the user's instructions. \\
You are provided with an evaluation rubric, but you must exercise \textbf{vigilance}. The provided rubric may be flawed, incomplete, or fundamentally incorrect (e.g., it might ignore the user's core intent or guide you toward a wrong conclusion). \\
You must not treat the provided rubric as absolute truth if it contradicts the user's instructions or logical reasoning. \\
\\
First, conduct a deep reasoning process inside \texttt{<analyze>} and \texttt{</analyze>} tags. Your reasoning must follow these steps: \\
\\
1. Rubric Validity Check \& Ideal Answer Formulation: \\
\quad - Analyze the User Question carefully. What is the core intent? \\
\quad - Formulate an \textbf{Ideal Answer} in your mind based on the User Question. What must a correct response contain? \\
\quad - Evaluate the provided Rubric. Does it align with the User Question and your Ideal Answer? \\
\quad - Determine if the rubric is \textbf{helpful} or \textbf{misleading}. \\
\quad\quad - If \textbf{helpful}: The rubric correctly captures the user's intent and logic. Use it as is. \\
\quad\quad - If \textbf{misleading}: The rubric contains errors, misses key constraints, or leads to incorrect evaluations. \textbf{You must explicitly discard the flawed parts and define your own correct rubric/criteria based on the User Question.} \\
\\
2. Step-by-Step Evaluation: \\
\quad - Compare Assistant A and Assistant B against the \textit{valid} criteria (either the provided reliable rubric or your newly defined correct rubric). \\
\quad - Be explicit: Which assistant matches the Ideal Answer better? \\
\\
3. Comparison: Based on the valid criteria, determine which assistant provided the superior response. \\
\\
Be explicit in your thought process. Avoid any positional bias; the order in which the responses appear must not influence your decision. Do not let response length or the assistants' names sway your judgment. \\
\\
\textbf{Output Format:} \\
After your reasoning, you must output \textbf{exactly and only} the following tags in this order, with no additional text: \\
1) \texttt{<analyze> ... </analyze>} \\
2) \texttt{<rubric>helpful</rubric>} \textbf{or} \texttt{<rubric>misleading</rubric>} \\
3) \texttt{<answer>A</answer>} \textbf{or} \texttt{<answer>B</answer>} \\
\\
Guidance for \texttt{<rubric>}: \\
- helpful: Choose this if the provided rubric is logical, accurate, and correctly guides the evaluation of the user's prompt. \\
- misleading: Choose this if the provided rubric is flawed, off the point, or would lead to selecting the wrong response if followed strictly. \\
\\
Below are the user's question, the two responses, and the evaluation rubric: \\
\\
{[}User Question{]} \\
\{prompt\} \\
\\
{[}The Start of Assistant A's Answer{]} \\
\{response\_a\} \\
{[}The End of Assistant A's Answer{]} \\
\\
{[}The Start of Assistant B's Answer{]} \\
\{response\_b\} \\
{[}The End of Assistant B's Answer{]} \\
\\
{[}The Start of RUBRIC{]} \\
\{criteria\} \\
{[}The End of RUBRIC{]} \\
\bottomrule
\end{tabular}
\caption{Prompt template for rubric-augmented verification used by $V_\theta$.}
\label{fig:prompt_rubric_aug}
\end{figure*}

\subsection{Rubric Quality Evaluation Prompt}
\label{app:prompts:quality_eval}

Figure~\ref{fig:prompt_quality_eval} contains the prompt used for GPT-5 to evaluate rubric quality on a 1--5 scale in Section~\ref{subsec:prelim_exp2} and Section~\ref{subsec:analysis_generator}.

\begin{figure*}[t]
\small
\begin{tabular}{p{0.95\linewidth}}
\toprule
\textbf{Rubric Quality Evaluation Prompt} \\
\midrule
You are an expert evaluator assessing the quality of auto-generated rubrics for comparing two AI assistant responses. \\
\\
\textbf{Your Task:} \\
Evaluate the generated rubric (including its reasoning in \texttt{<analyze>} tags and the criteria/rubric questions) on a 1-5 scale based on how effectively it enables accurate quality comparison between two responses. \\
\\
\textbf{Input:} \\
- User Question \\
- Assistant A's Answer \\
- Assistant B's Answer \\
- Generated Rubric (analysis + criteria + rubric questions) \\
\\
\textbf{Evaluation Criteria (1-5 Scale):} \\
\\
\textbf{Score 1 - Misleading/Harmful:} \\
The rubric would lead evaluators to incorrect judgments. Rubric questions are irrelevant to the actual quality differences, focus on superficial aspects while ignoring critical flaws, or are biased toward the objectively worse response. Analysis is absent, incoherent, or contradicts the generated rubrics. \\
\\
\textbf{Score 2 - Poor:} \\
The rubric fails to capture key differences. Analysis is shallow, rubric questions are too vague to be actionable (e.g., ``Is the response good?'') or non-discriminative (would yield identical answers for both A and B). \\
\\
\textbf{Score 3 - Acceptable:} \\
The rubric identifies some valid differences but misses important ones. Analysis follows the structure but lacks depth. Rubric questions are answerable but either too generic or only partially aligned with the core quality gap between responses. \\
\\
\textbf{Score 4 - Good:} \\
The rubric effectively highlights the main quality differences. Analysis is thorough and logically derives the criteria. Rubric questions are specific, discriminative, and appropriately weighted toward what matters most for the user's intent. Minor gaps or suboptimal criterion selection. \\
\\
\textbf{Score 5 - Excellent:} \\
The rubric precisely targets the decisive quality differences. Analysis demonstrates clear reasoning through all steps with accurate characterization of both responses. Rubric questions are highly specific, directly answerable, and would reliably guide any evaluator to the correct quality judgment. Criteria selection is optimal. \\
\\
\textbf{Output Format:} \\
\texttt{<evaluation>} \\
{[}Brief justification highlighting: (1) analysis quality, (2) rubric specificity/discriminativeness, (3) alignment with actual response differences{]} \\
\texttt{</evaluation>} \\
\texttt{<score>}{[}1-5{]}\texttt{</score>} \\
\\
--- \\
\\
{[}User Question{]} \\
\{question\} \\
\\
{[}Assistant A's Answer{]} \\
\{response\_a\} \\
\\
{[}Assistant B's Answer{]} \\
\{response\_b\} \\
\\
{[}Generated Rubric{]} \\
\{rubric\} \\
\bottomrule
\end{tabular}
\caption{Prompt template for rubric quality evaluation using GPT-5.}
\label{fig:prompt_quality_eval}
\end{figure*}

\section{Implementation Details}
\label{app:implementation}

\paragraph{Training Hyperparameters}

Table~\ref{tab:hyperparams_grpo} and Table~\ref{tab:hyperparams_dpo} summarize the hyperparameters used for training the verifier (GRPO) and the rubric generator (DPO), respectively.
The GRPO hyperparameters were used consistently for both Reasoning RM and C2's RL training, except for the number of epochs: Reasoning RM was trained for 3 epochs, while C2 was trained for 1 epoch.
This difference accounts for the fact that C2's training data is augmented with rubric-augmented tasks (one helpful and one misleading rubric per example), resulting in approximately 3$\times$ the data size of Reasoning RM's rubric-free data, ensuring comparable training compute across methods.

For downstream RLHF experiments (Section~\ref{subsec:exp_rlhf}), we use the same DPO hyperparameters as Table~\ref{tab:hyperparams_dpo} to fine-tune the base model on the preference pairs constructed by C2 and Reasoning RM.
For Qwen3-8B, we set \texttt{enable\_thinking=False} during response sampling for RLHF and when evaluating the optimized policy on AlpacaEval 2.0 and Arena-Hard, as keeping it enabled caused significant performance degradation.

We used trl \cite{vonwerra2022trl} for both DPO and GRPO training, and vLLM \cite{kwon2023efficient} for inference.
All experiments were conducted on 8 NVIDIA A100 80GB GPUs.

\begin{table}[h]
\centering
\small
\begin{tabular}{lcc}
\toprule
\textbf{Hyperparameter} & \textbf{Tulu3-8B} & \textbf{Qwen3-8B} \\
\midrule
Learning rate & 5e-7 & 5e-7 \\
Batch size & 64 & 64 \\
Mini-batch size & 32 & 32 \\
Number of epochs & \multicolumn{2}{c}{3 (Reasoning RM) / 1 (C2)} \\
KL loss coefficient & 0.01 & 0.01 \\
Number of rollouts & 8 & 8 \\
Sampling temperature & 1.0 & 0.6 \\
Max prompt length & 8192 & 8192 \\
Max response length & 2048 & 2048 \\
LR scheduler & Linear & Linear \\
Warmup ratio & 0.1 & 0.1 \\
\bottomrule
\end{tabular}
\caption{Hyperparameters for GRPO training (Verifier).}
\label{tab:hyperparams_grpo}
\end{table}

\begin{table}[h]
\centering
\small
\begin{tabular}{lc}
\toprule
\textbf{Hyperparameter} & \textbf{Tulu3-8B / Qwen3-8B} \\
\midrule
Learning rate & 5e-7 \\
Batch size & 64 \\
Number of epochs & 3 \\
$\beta$ & 0.1 \\
Max sequence length & 4096 \\
Optimizer & AdamW \\
Weight decay & 0.01 \\
Warmup ratio & 0.1 \\
\bottomrule
\end{tabular}
\caption{Hyperparameters for DPO training (Rubric Generator). The same hyperparameters are used for both Tulu3-8B and Qwen3-8B.}
\label{tab:hyperparams_dpo}
\end{table}

\paragraph{Inference Token Consumption}
Section~\ref{subsec:analysis_compute} compares C2 against compute-matched Reasoning RM baselines. 
On RewardBench, the average number of generated tokens per example is 803 for Reasoning RM and 1,862 for C2 with Tulu3-8B-SFT (2.3$\times$), and 1,018 for Reasoning RM and 2,465 for C2 with Qwen3-8B (2.4$\times$). 
This overhead arises from rubric generation and the potential retry mechanism when rubrics are flagged as misleading.

\paragraph{Rubric Pair Sampling}
When synthesizing helpful and misleading rubric pairs (Section~\ref{subsec:c2_rubric_synthesis}), sampling $K=16$ rubric candidates from $M_g$ does not always yield at least one rubric for each of $\mathcal{R}^+$ and $\mathcal{R}^-$. 
For such examples, we repeated the sampling procedure up to 5 additional times until a helpful and misleading rubric pair was obtained, and discarded the example only if no pair could be formed after these retries.

\section{Reward Function Details}
\label{app:reward}

\subsection{Reward Components}
\label{app:reward:components}

We decompose the reward function into three binary components, each yielding $+1$ on success and $-1$ otherwise:

\paragraph{Format Reward ($R_f$)} Checks whether the output follows the required structure. For rubric-free verification:
\begin{equation}
R_f = 
\begin{cases}
+1 & \text{if format is valid} \\
-1 & \text{otherwise}
\end{cases}
\end{equation}
where the valid format is \texttt{<analyze>...\allowbreak <answer>...}.

For rubric-augmented verification:
\begin{equation}
R_f = 
\begin{cases}
+1 & \text{if output matches the required format} \\
-1 & \text{otherwise}
\end{cases}
\end{equation}
where the required format is \texttt{<analyze>...<rubric>...<answer>...}.

\paragraph{Preference Reward ($R_p$)} Checks whether the predicted preference matches the gold label:
\begin{equation}
R_p = 
\begin{cases}
+1 & \text{if } \hat{l} = l \\
-1 & \text{otherwise}
\end{cases}
\end{equation}

\paragraph{Rubric Reward ($R_r$)} For rubric-augmented tasks, checks whether the rubric assessment matches the synthesized label:
\begin{equation}
R_r = 
\begin{cases}
+1 & \text{if } q = q^* \\
-1 & \text{otherwise}
\end{cases}
\end{equation}
where $q^*$ is \texttt{helpful} for $r^+$ and \texttt{misleading} for $r^-$.

\subsection{Reward Weights}
\label{app:reward:weights}
The total reward is computed as a weighted sum of the components:
\paragraph{Rubric-free task:}
\begin{equation}
R = w_f \cdot R_f + w_p \cdot R_p
\end{equation}

\paragraph{Rubric-augmented task:}
\begin{equation}
R = w_f \cdot R_f + w_p \cdot R_p + w_r \cdot R_r
\end{equation}

We selected reward weights based on performance on a held-out validation set of 500 examples from UltraFeedback.
For Reasoning RM, we searched over $(w_p, w_f) \in \{(0.9, 0.1), (0.8, 0.2), (0.7, 0.3)\}$ and selected $(0.8, 0.2)$ for Tulu3-8B and $(0.9, 0.1)$ for Qwen3-8B.
For C2, we fixed $w_f = 0.1$ and searched over $w_p \in \{0.7, 0.6, 0.5, 0.4\}$ with $w_r = 0.9 - w_p$, selecting $(w_p, w_r, w_f) = (0.6, 0.3, 0.1)$ for both models.

\section{Additional RLHF Experiments}
\label{app:additional_rlhf}

To validate that C2's improvements in reward modeling consistently transfer to downstream performance beyond DPO (Section~\ref{subsec:exp_rlhf}), we conduct additional experiments using best-of-N selection and rejection sampling with Qwen3-8B.

\subsection{Best-of-N Selection}
\label{app:additional_rlhf:bon}

We sample $N$ candidate responses from Qwen3-8B (\texttt{enable\_thinking} set to false) and use C2 and Reasoning RM (Qwen3-8B as base) to select the best response across five benchmarks spanning reasoning, instruction following, and open-ended generation.

\begin{table}[h]
\centering
\small
\setlength{\tabcolsep}{5pt}
\begin{tabular}{llccc}
\toprule
\textbf{Benchmark} & \textbf{Method} & $N{=}4$ & $N{=}8$ & $N{=}16$ \\
\midrule
\multirow{2}{*}{GPQA-Diamond} & RRM & 41.4 & 43.4 & 44.4 \\
 & C2 & \textbf{43.9} & \textbf{46.5} & \textbf{47.5} \\
\midrule
\multirow{2}{*}{MATH500} & RRM & 88.2 & 89.8 & 90.2 \\
 & C2 & \textbf{90.6} & \textbf{92.0} & \textbf{92.6} \\
\midrule
\multirow{2}{*}{LogiQA} & RRM & 73.0 & 73.3 & 73.3 \\
 & C2 & \textbf{73.4} & \textbf{73.6} & \textbf{73.7} \\
\midrule
\multirow{2}{*}{IFEval (strict)} & RRM & 85.0 & 85.7 & 86.1 \\
 & C2 & \textbf{86.7} & \textbf{87.2} & \textbf{87.8} \\
\midrule
\multirow{2}{*}{Arena-Hard} & RRM & 72.1 & 73.2 & 73.8 \\
 & C2 & \textbf{74.8} & \textbf{75.9} & \textbf{76.8} \\
\bottomrule
\end{tabular}
\caption{Best-of-N selection results (\%) with Qwen3-8B. C2 consistently outperforms Reasoning RM (RRM) across all benchmarks and values of $N$, with gains particularly pronounced on GPQA-Diamond (+3.1 at $N{=}16$) and MATH500 (+2.4 at $N{=}16$).}
\label{tab:bon_results}
\end{table}

Table~\ref{tab:bon_results} shows that C2 consistently outperforms Reasoning RM across all benchmarks and all values of $N$.
The gains are particularly pronounced on challenging reasoning tasks, with +3.1 on GPQA-Diamond and +2.4 on MATH500 at $N{=}16$.

\subsection{Rejection Sampling}
\label{app:additional_rlhf:rejection}

Following the same protocol as Section~\ref{subsec:exp_rlhf}, we sample 8 candidate responses per prompt and use each reward model (C2 or Reasoning RM) to select the best response, then fine-tune Qwen3-8B on the selected responses via SFT.

\begin{table}[h]
\centering
\small
\begin{tabular}{lccc}
\toprule
& \multicolumn{2}{c}{\textbf{AlpacaEval 2.0}} & \textbf{Arena-Hard} \\
\cmidrule(lr){2-3} \cmidrule(lr){4-4}
\textbf{Method} & WR & LC & WR \\
\midrule
Reasoning RM & 39.9 & 39.3 & 70.4 \\
C2 (Ours) & \textbf{42.1} & \textbf{41.3} & \textbf{72.6} \\
\bottomrule
\end{tabular}
\caption{Rejection sampling results (\%) with Qwen3-8B. C2 yields +2.0 LC win rate on AlpacaEval 2.0 and +2.2 on Arena-Hard.}
\label{tab:rejection_sampling_results}
\end{table}

Table~\ref{tab:rejection_sampling_results} shows that C2 yields +2.0 LC win rate on AlpacaEval 2.0 and +2.2 on Arena-Hard, confirming that the gains transfer to rejection sampling as well.

\section{Inference Latency}
\label{app:latency}

Table~\ref{tab:latency} reports per-example inference latency measured on RewardBench using a single NVIDIA A100 80GB GPU with vLLM.

\begin{table}[h]
\centering
\small
\begin{tabular}{lcc}
\toprule
\textbf{Model} & \textbf{Reasoning RM} & \textbf{C2} \\
\midrule
Tulu3-8B-SFT & 1,551 $\pm$ 766 ms & 3,628 $\pm$ 1,801 ms \\
Qwen3-8B & 1,989 $\pm$ 981 ms & 4,834 $\pm$ 2,393 ms \\
\bottomrule
\end{tabular}
\caption{Per-example inference latency (mean $\pm$ std) on RewardBench using a single A100 GPU. C2 is approximately 2.3--2.4$\times$ slower than Reasoning RM due to rubric generation and the potential retry mechanism.}
\label{tab:latency}
\end{table}

C2 is approximately 2.3--2.4$\times$ slower than Reasoning RM, consistent with the token consumption ratio reported in Appendix~\ref{app:implementation}.
This overhead arises from rubric generation and the potential retry mechanism when rubrics are flagged as misleading.
While C2 incurs higher latency, the compute-matched experiments in Section~\ref{subsec:analysis_compute} confirm that C2's gains stem from its cooperative-critical design rather than from additional computation alone.

\section{Qualitative Analysis}
\label{app:qualitative}

\subsection{Rubric Error Analysis}
\label{app:qualitative:error_analysis}

To better understand how C2's training affects rubric quality beyond aggregate scores (Section~\ref{subsec:analysis_generator}), we conduct a structured error analysis comparing rubrics from the base model and the C2 generator.

\paragraph{Setup}
We sample 80 rubrics each from the base model and the C2 rubric generator (Qwen3-8B as the base model) on examples from RM-Bench.
All rubrics are annotated by the authors in a blind setting, where the annotator did not know whether each rubric was generated by the base model or by C2.
We use the following taxonomy of failure modes, where multiple labels can apply to a single rubric:

\begin{itemize}
    \item \textbf{Missing key constraints}: The rubric omits important evaluation criteria required by the prompt.
    \item \textbf{Irrelevant criteria}: The rubric includes one or more criteria unrelated to the prompt's requirements.
    \item \textbf{Contradictory}: The rubric contains criteria that are mutually difficult to satisfy or internally inconsistent.
    \item \textbf{Ambiguous}: The rubric relies solely on abstract criteria (e.g., ``Is it accurate?'') without specific, actionable evaluation questions.
    \item \textbf{Over-constrained}: The rubric imposes one or more requirements beyond the scope of the prompt.
\end{itemize}
Rubrics with no applicable error labels are assigned a \textit{no error} label.

\paragraph{Results}

\begin{table}[h]
\centering
\small
\begin{tabular}{lcc}
\toprule
\textbf{Error Type} & \textbf{Base Model} & \textbf{C2} \\
\midrule
Missing key constraints & 17 (21.3\%) & 8 (10.0\%) \\
Irrelevant criteria & 21 (26.3\%) & 10 (12.5\%) \\
Contradictory & 3 (3.8\%) & 3 (3.8\%) \\
Ambiguous & 27 (33.8\%) & 4 (5.0\%) \\
Over-constrained & 10 (12.5\%) & 15 (18.8\%) \\
\midrule
No error & 28 (35.0\%) & 42 (52.5\%) \\
\bottomrule
\end{tabular}
\caption{Error analysis of rubrics generated by the base model vs.\ C2 generator (Qwen3-8B). Counts and percentages are shown for each error type over 80 rubrics per model. Multiple error labels can apply to a single rubric.}
\label{tab:error_analysis}
\end{table}

Table~\ref{tab:error_analysis} presents the results.
C2 raises the error-free rate from 35.0\% to 52.5\% (a 50\% relative improvement).
The largest reductions are in irrelevant criteria (26.3\%$\to$12.5\%) and ambiguous rubrics (33.8\%$\to$5.0\%), indicating that C2 produces more focused and discriminative criteria.
Missing key constraints also decrease substantially (21.3\%$\to$10.0\%).
A minor side effect is a slight increase in over-constrained rubrics (12.5\%$\to$18.8\%), which appears to be a consequence of training toward more discriminative rubrics.
The generator occasionally introduces requirements beyond the prompt's scope in an effort to sharpen the distinction between responses.
Overall, these patterns confirm that C2's contrastive training produces clear, prompt-focused rubrics rather than semantically uninterpretable artifacts.

\subsection{Examples of High-Quality and Low-Quality Rubrics}
\label{app:qualitative:quality}

We present examples of high-quality and low-quality rubrics generated for the same prompt: Figure~\ref{fig:qual_example1} shows examples from Qwen3-8B and Figure~\ref{fig:qual_example2} from Tulu3-8B-SFT.

\begin{figure*}[t]
\small
\begin{tabular}{p{0.47\linewidth} p{0.47\linewidth}}
\toprule
\multicolumn{2}{p{0.96\linewidth}}{\textbf{Prompt:} Implement a program to find the common elements in two arrays without using any extra data structures.} \\
\midrule
\textbf{High-Quality Rubric (Score: 4)} & \textbf{Low-Quality Rubric (Score: 2)} \\
\midrule
\texttt{<analyze>} \textbf{Intent:} The user wants a program to find common elements in two arrays \textbf{without using any extra data structures}. The core intent is to solve the problem efficiently while adhering to the constraint of no additional memory allocation. \textbf{Ideal Answer:} A model answer should use a method that does not require extra memory (e.g., nested loops for unsorted arrays or two-pointer technique for sorted arrays), avoid using lists, sets, or dictionaries to store intermediate results, and provide a working code example. \texttt{</analyze>}

\vspace{1mm}
\texttt{<criteria\_1>} Instruction-following \texttt{</criteria\_1>}
\newline
\texttt{<rubric\_1>} Does the response avoid using any extra data structures (e.g., lists, sets, dictionaries) to store intermediate or final results, strictly adhering to the ``no extra data structures'' constraint? \texttt{</rubric\_1>}

\vspace{1mm}
\texttt{<criteria\_2>} Helpfulness \texttt{</criteria\_2>}
\newline
\texttt{<rubric\_2>} Does the response provide a working code example that correctly identifies common elements while following the constraint? \texttt{</rubric\_2>}

\vspace{1mm}
\texttt{<criteria\_3>} Completeness \texttt{</criteria\_3>}
\newline
\texttt{<rubric\_3>} Does the response include all necessary details (e.g., algorithm logic, code, explanation) to solve the problem? \texttt{</rubric\_3>}

&

\texttt{<analyze>} The user's intent is to implement a program that finds common elements without extra data structures. Assistant B offers a nested loop implementation but uses a list to store results, which might violate the constraint. However, the result list is necessary for output, so it may be acceptable. \texttt{</analyze>}

\vspace{1mm}
\texttt{<criteria\_1>} Helpfulness \texttt{</criteria\_1>}
\newline
\texttt{<rubric\_1>} Does the response provide a working implementation that correctly identifies common elements? \texttt{</rubric\_1>}

\vspace{1mm}
\texttt{<criteria\_2>} Completeness \texttt{</criteria\_2>}
\newline
\texttt{<rubric\_2>} Does the response include all necessary code, including the use of a result list to store common elements? \texttt{</rubric\_2>}

\vspace{1mm}
\texttt{<criteria\_3>} Instruction-following \texttt{</criteria\_3>}
\newline
\texttt{<rubric\_3>} Does the response avoid using structures like lists for storing results, even if they are required for output? \texttt{</rubric\_3>}

\\
\bottomrule
\end{tabular}
\caption{High-quality vs. low-quality rubrics for implementing common element detection without extra data structures (Qwen3-8B). The high-quality rubric clearly defines the ``no extra data structures'' constraint and derives consistent evaluation criteria from it. The low-quality rubric shows ambiguous interpretation of the constraint, stating that violating it ``may be acceptable,'' leading to contradictory criteria that simultaneously require and forbid using a result list.}
\label{fig:qual_example1}
\end{figure*}

\begin{figure*}[t]
\small
\begin{tabular}{p{0.47\linewidth} p{0.47\linewidth}}
\toprule
\multicolumn{2}{p{0.96\linewidth}}{\textbf{Prompt:} I've put a plate on top of an apple, and take the plate from kitchen to dinning room. Where is the apple now} \\
\midrule
\textbf{High-Quality Rubric (Score: 4)} & \textbf{Low-Quality Rubric (Score: 1)} \\
\midrule
\texttt{<analyze>} \textbf{Intent:} The user wants to know the location of the apple after it has been moved along with a plate from the kitchen to the dining room. \textbf{Ideal Answer:} A model answer should clearly state that the apple remains in the kitchen while the plate is moved to the dining room. The apple's location is dictated by the plate's position when it was initially placed on top of it, rather than by subsequent movements of the plate. \texttt{</analyze>}

\vspace{1mm}
\texttt{<criteria\_1>} Completeness \texttt{</criteria\_1>}
\newline
\texttt{<rubric\_1>} Does the response accurately describe the apple's final location and mention that the apple did not move with the plate? \texttt{</rubric\_1>}

\vspace{1mm}
\texttt{<criteria\_2>} Safety \texttt{</criteria\_2>}
\newline
\texttt{<rubric\_2>} Does the response avoid any content that may imply the apple is in the dining room or mislead the user? \texttt{</rubric\_2>}

\vspace{1mm}
\texttt{<criteria\_3>} Helpfulness \texttt{</criteria\_3>}
\newline
\texttt{<rubric\_3>} Does the response provide a clear explanation that directly addresses the user's question about the apple's location? \texttt{</rubric\_3>}

&

\texttt{<analyze>} \textbf{Intent:} The user is asking about the location of the apple in relation to the plate after moving the plate. \textbf{Ideal Answer:} A model answer should specify that the apple remains in the kitchen, directly under the plate, since the apple was concealed underneath the plate and did not move. \texttt{</analyze>}

\vspace{1mm}
\texttt{<criteria\_1>} Helpfulness \texttt{</criteria\_1>}
\newline
\texttt{<rubric\_1>} Does the response accurately explain the final location of the apple? \texttt{</rubric\_1>}

\vspace{1mm}
\texttt{<criteria\_2>} Completeness \texttt{</criteria\_2>}
\newline
\texttt{<rubric\_2>} Does the response provide a full explanation including initial and final positions of both objects? \texttt{</rubric\_2>}

\vspace{1mm}
\texttt{<criteria\_3>} Safety \texttt{</criteria\_3>}
\newline
\texttt{<rubric\_3>} Does the response avoid introducing any dangerous, harmful, or false information about relative positions? \texttt{</rubric\_3>}

\\
\bottomrule
\end{tabular}
\caption{High-quality vs. low-quality rubrics for a spatial reasoning question (Tulu3-8B-SFT). The high-quality rubric correctly interprets ``plate on top of apple'' and reasons that moving the plate leaves the apple stationary in the kitchen. The low-quality rubric misunderstands the physical configuration (claiming the apple is ``concealed underneath the plate''), and inappropriately applies the Safety criterion to object positions rather than harmful content.}
\label{fig:qual_example2}
\end{figure*}

\subsection{Rubric Generation Comparison Across Models}
\label{app:qualitative:comparison}

We compare rubrics generated by the base model, C2 generator (after DPO training), and a larger model from the same family for the same prompt: Figure~\ref{fig:comparison_tulu3} shows examples from the Tulu3 family and Figure~\ref{fig:comparison_qwen3} from the Qwen3 family.

\begin{figure*}[t]
\small
\centering

\begin{tabular}{p{0.95\linewidth}}
\toprule
\multicolumn{1}{p{0.95\linewidth}}{\textbf{Prompt:} Let $k$ and $m$ be real numbers, and suppose that the roots of the equation $x^3 - 7x^2 + kx - m = 0$ are three distinct positive integers. Compute $k + m.$} \\
\midrule
\textbf{Assistant A (Chosen Response):} \\
\midrule
Given the polynomial $x^3 - 7x^2 + kx - m = 0$ with roots $a$, $b$, $c$ as distinct positive integers, by Vieta's formulas: $a + b + c = 7$, $ab + bc + ca = k$, $abc = m$. Since $a$, $b$, $c$ are distinct positive integers summing to 7, the only solution is $\{1, 2, 4\}$. Thus: $k = 1 \cdot 2 + 2 \cdot 4 + 4 \cdot 1 = 14$, $m = 1 \cdot 2 \cdot 4 = 8$. Therefore, $k + m = 14 + 8 = \mathbf{22}$.
\\
\midrule
\textbf{Assistant B (Rejected Response):} \\
\midrule
Given the polynomial with roots as distinct positive integers, by Vieta's formulas: $a + b + c = 7$. The smallest distinct positive integers are $1, 2, 3$. Using these: $k = 1 \cdot 2 + 2 \cdot 3 + 3 \cdot 1 = 11$, $m = 1 \cdot 2 \cdot 3 = 6$. Therefore, $k + m = 11 + 6 = \mathbf{17}$.
\\
\midrule
\textbf{Base Model} (Tulu3-8B-SFT, Score: 2) \\
\midrule
\texttt{<analyze>}

Intent: The User Question seeks to find the values of $k$ and $m$ in the polynomial equation given that the roots are three distinct positive integers. The goal is to compute $k + m$.

Comparison: Assistant A correctly identifies the roots as 1, 2, and 4 and computes $k + m$ as 22. Assistant B correctly identifies the roots as 1, 2, and 3 and computes $k + m$ as 17. Assistant A's answer is more accurate because it correctly matches the polynomial's roots with the given constraints.

\texttt{</analyze>}

\texttt{<criteria\_1>} Completeness \texttt{</criteria\_1>}
\texttt{<rubric\_1>} Does the answer include all necessary steps to derive the values of $k$ and $m$? \texttt{</rubric\_1>}

\texttt{<criteria\_2>} Helpfulness \texttt{</criteria\_2>}
\texttt{<rubric\_2>} Does the answer directly address the user's intent and provide a correct and relevant solution? \texttt{</rubric\_2>}
\\
\midrule
\textbf{C2 Generator} (Tulu3-8B-SFT + DPO, Score: 4) \\
\midrule
\texttt{<analyze>}

Intent: The core point of the user question is to find the values of $k$ and $m$ such that the roots of the polynomial are three distinct positive integers, and then compute $k + m$.

Ideal Answer: The ideal answer should correctly identify the roots that are distinct positive integers summing to 7, use Vieta's formulas to calculate $k$ and $m$, and compute $k + m$ accurately.

\texttt{</analyze>}

\texttt{<criteria\_1>} Helpfulness \texttt{</criteria\_1>}
\texttt{<rubric\_1>} Does the answer directly address the user's question of finding $k$ and $m$ and computing $k + m$? \texttt{</rubric\_1>}

\texttt{<criteria\_2>} Completeness \texttt{</criteria\_2>}
\texttt{<rubric\_2>} Does the answer correctly identify the roots that satisfy the condition $a + b + c = 7$ and use Vieta's formulas to compute $k$ and $m$? \texttt{</rubric\_2>}

\texttt{<criteria\_3>} Accuracy in calculations \texttt{</criteria\_3>}
\texttt{<rubric\_3>} Are the values of $k$ and $m$ calculated accurately, and does the sum $k + m$ match the correct answer? \texttt{</rubric\_3>}
\\
\midrule
\textbf{Larger Model} (Tulu3-70B, Score: 5) \\
\midrule
\texttt{<analyze>}

Intent: The core point of the User Question is to determine the values of $k$ and $m$ given that the roots of the polynomial are three distinct positive integers and then compute $k + m$.

Ideal Answer: The ideal answer should correctly identify the roots as 1, 2, and 4, calculate $k$ and $m$ based on these roots, and then compute $k + m$.

\texttt{</analyze>}

\texttt{<criteria\_1>} Helpfulness \texttt{</criteria\_1>}
\texttt{<rubric\_1>} Does the answer correctly identify the roots of the polynomial as distinct positive integers that sum to 7? \texttt{</rubric\_1>}

\texttt{<criteria\_2>} Completeness \texttt{</criteria\_2>}
\texttt{<rubric\_2>} Does the answer correctly compute $k$ and $m$ based on the identified roots? \texttt{</rubric\_2>}

\texttt{<criteria\_3>} Instruction-following \texttt{</criteria\_3>}
\texttt{<rubric\_3>} Does the answer accurately compute $k + m$? \texttt{</rubric\_3>}
\\
\bottomrule
\end{tabular}

\caption{Rubric comparison across Tulu3 family models. The base model produces generic rubrics (``Completeness'' and ``Helpfulness'') that fail to target the critical constraint: roots must sum to 7. The C2 generator explicitly identifies this constraint in its criteria, enabling accurate discrimination between correct (roots 1, 2, 4) and incorrect (roots 1, 2, 3) solutions.}
\label{fig:comparison_tulu3}
\end{figure*}

\begin{figure*}[t]
\small
\centering

\begin{tabular}{p{0.95\linewidth}}
\toprule
\multicolumn{1}{p{0.95\linewidth}}{\textbf{Prompt:} The greatest common divisor of positive integers $m$ and $n$ is 8. The least common multiple of $m$ and $n$ is 112. What is the least possible value of $m+n$?} \\
\midrule
\textbf{Assistant A (Chosen Response):} \\
\midrule
Using $\gcd(m,n) \cdot \text{lcm}(m,n) = m \cdot n$, we get $8 \cdot 112 = 896$, so $m \cdot n = 896$. Let $m = 8a$ and $n = 8b$ where $\gcd(a,b) = 1$. Then $64ab = 896$, giving $ab = 14$. The coprime pairs $(a,b)$ with product 14 are: $(1, 14)$ and $(2, 7)$. For $(1, 14)$: $m + n = 8 + 112 = 120$. For $(2, 7)$: $m + n = 16 + 56 = 72$. The minimum is $\mathbf{72}$.
\\
\midrule
\textbf{Assistant B (Rejected Response):} \\
\midrule
Using $\gcd(m,n) \cdot \text{lcm}(m,n) = m \cdot n$, we get $m \cdot n = 896$. Let $m = 8a$ and $n = 8b$ where $\gcd(a,b) = 1$. Then $ab = 14$. The pairs $(a,b)$ satisfying this are $(1, 14)$ and $(14, 1)$. For both cases: $m + n = 8 + 112 = 120$. Therefore, the minimum is $\mathbf{120}$.
\\
\midrule
\textbf{Base Model} (Qwen3-8B, Score: 3) \\
\midrule
\texttt{<analyze>}

Intent: The user wants the minimal sum of two numbers with specific GCD and LCM. The key here is to use the relationship between GCD and LCM.

Ideal Answer: Must enumerate all coprime pairs of $a$ and $b$ that multiply to 14, compute $m + n$ for each pair and select the minimum.

\texttt{</analyze>}

\texttt{<criteria\_1>} Completeness \texttt{</criteria\_1>}
\texttt{<rubric\_1>} Does the answer include all valid coprime pairs $(a, b)$ such that $ab = 14$, including $(1, 14), (2, 7), (7, 2), (14, 1)$? \texttt{</rubric\_1>}

\texttt{<criteria\_2>} Helpfulness \texttt{</criteria\_2>}
\texttt{<rubric\_2>} Does the answer correctly identify the minimal possible value of $m + n$ as 72, or does it provide an incorrect value (e.g., 120)? \texttt{</rubric\_2>}

\texttt{<criteria\_3>} Instruction-following \texttt{</criteria\_3>}
\texttt{<rubric\_3>} Does the answer adhere to the mathematical principles and logical steps required to solve the problem? \texttt{</rubric\_3>}
\\
\midrule
\textbf{C2 Generator} (Qwen3-8B + DPO, Score: 4) \\
\midrule
\texttt{<analyze>}

Intent: The user seeks the minimal sum of two integers $m$ and $n$ given their GCD is 8 and LCM is 112.

Ideal Answer: A complete answer must use the formula $\text{GCD} \times \text{LCM} = m \times n$, factor $m = 8a$, $n = 8b$, enumerate all coprime pairs, and select the minimum.

\texttt{</analyze>}

\texttt{<criteria\_1>} Completeness \texttt{</criteria\_1>}
\texttt{<rubric\_1>} Does the answer include all valid coprime factor pairs of $ab = 14$ (e.g., $(1,14)$ and $(2,7)$) and compute the corresponding $m+n$ values? \texttt{</rubric\_1>}

\texttt{<criteria\_2>} Helpfulness \texttt{</criteria\_2>}
\texttt{<rubric\_2>} Does the answer correctly identify the least possible value of $m+n$ as 72, or does it provide an incorrect minimal sum (e.g., 120)? \texttt{</rubric\_2>}
\\
\midrule
\textbf{Larger Model} (Qwen3-32B, Score: 4) \\
\midrule
\texttt{<analyze>}

Intent: The user wants to find the minimal sum of two positive integers $m$ and $n$ given their GCD is 8 and LCM is 112.

Ideal Answer: Must correctly enumerate all coprime pairs and select the minimum sum.

\texttt{</analyze>}

\texttt{<criteria\_1>} Completeness \texttt{</criteria\_1>}
\texttt{<rubric\_1>} Does the answer enumerate all coprime integer pairs $(a, b)$ such that $ab = 14$? \texttt{</rubric\_1>}

\texttt{<criteria\_2>} Helpfulness \texttt{</criteria\_2>}
\texttt{<rubric\_2>} Does the answer correctly identify the minimum possible value of $m + n$ based on the derived pairs? \texttt{</rubric\_2>}

\texttt{<criteria\_3>} Instruction-following \texttt{</criteria\_3>}
\texttt{<rubric\_3>} Does the answer adhere to the mathematical process of factoring $m$ and $n$ as multiples of their GCD? \texttt{</rubric\_3>}
\\
\bottomrule
\end{tabular}

\caption{Rubric comparison across Qwen3 family models. All three models correctly identify the key issue---the missing $(2,7)$ coprime pair---but differ in rubric specificity. The C2 generator produces the most focused criteria by directly asking whether all coprime pairs are enumerated and whether the correct minimum (72) is identified, without redundant criteria.}
\label{fig:comparison_qwen3}
\end{figure*}

\subsection{Verifier Reasoning Examples}
\label{app:qualitative:reasoning}

We present examples illustrating how the C2 verifier reasons about rubric quality and makes preference judgments.
The verifier can correctly leverage helpful rubrics (Figure~\ref{fig:verifier_case1}), appropriately reject misleading rubrics (Figure~\ref{fig:verifier_case2}), but can also make errors, incorrectly dismissing helpful rubrics (Figure~\ref{fig:verifier_case3}) or trusting misleading ones (Figure~\ref{fig:verifier_case4}).
These cases demonstrate both the capabilities and limitations of critical verification.

%%%%%%%%%%%%%%%%%%%%%%%%%%%%%%%%%%%%%%%%%%%%%%%%%%%%%%%%%%%%%%%%%%%%%%%%%%%%%%%
% Case 1: Helpful rubric, verifier marks reliable, correct decision
%%%%%%%%%%%%%%%%%%%%%%%%%%%%%%%%%%%%%%%%%%%%%%%%%%%%%%%%%%%%%%%%%%%%%%%%%%%%%%%

\begin{figure*}[t]
\small
\centering
\begin{tabular}{p{0.95\linewidth}}
\toprule
\textbf{Case 1: Correctly Following a Helpful Rubric} \hfill \textit{(Qwen3-8B)} \\
\midrule
\textbf{Prompt:} How can I make bubble solution? \\
\midrule
\textbf{Response A:} \\
Create bubble solution by mixing water, vinegar, and optional glycerin or corn syrup, then use a homemade wand to blow bubbles for a fun activity. \\
\midrule
\textbf{Response B:} \\
Making your own bubble solution at home is a fun and easy activity for kids and adults alike. Here are a few simple recipes to get you started. Basic Bubble Solution Recipe. Ingredients: 6 cups of water, 1/2 cup of dish soap (like Dawn or Joy), 2 tablespoons of glycerin (optional, but helps create longer-lasting bubbles). Instructions: Mix the water and dish soap in a large container. If using glycerin, add it to the mixture and stir well. Let the solution sit for a few hours or overnight to allow the soap to dissolve fully and the bubbles to become more stable. Use a bubble wand or a homemade wand (like a wire hanger bent into a circle) to blow bubbles. Advanced Bubble Solution Recipe. Ingredients: 4 cups of water, 1/2 cup of dish soap, 1/4 cup of corn syrup (helps create larger bubbles), 1 tablespoon of baking powder (optional, for extra strength). Instructions: Combine the water, dish soap, and corn syrup in a large container. If using baking powder, add it to the mixture and stir until dissolved. Let the solution sit for a few hours or overnight. Use a bubble wand to blow bubbles. Homemade Bubble Wands. If you don't have a bubble wand, you can easily make one. Wire Hanger Wand: Straighten a wire hanger and bend one end into a circle. Twist the ends together to secure the circle. Yarn Wand: Wrap yarn around a wire hanger or a stick to create a fluffy wand that can catch more solution. Plastic Ring Wand: Use the ring from a milk jug or any other plastic ring. Dip it into the solution and blow through the center. Tips for Blowing Bubbles. Warm Solution: Warm water can help create better bubbles. Gentle Blowing: Blow gently to create larger, longer-lasting bubbles. Outdoor Play: Bubbles are best enjoyed outdoors where there is less chance of them popping quickly due to dust or drafts. Enjoy your bubble-blowing adventure! \\
\midrule
\textbf{Generated Rubric:} \\
The user is asking for a method to create bubble solution, likely for a fun activity. The core intent is to provide a clear, actionable recipe with ingredients and steps. A good answer should be practical, safe, and detailed enough for someone to replicate successfully.

The ideal answer would include: 1. Clear ingredients (e.g., water, dish soap, optional additives like glycerin/corn syrup). 2. Step-by-step instructions. 3. Optional variations or advanced techniques. 4. Safety notes (e.g., avoiding harmful substances).

Assistant B excels in Completeness and Helpfulness, covering all necessary aspects with accurate information. Assistant A is incomplete, lacks detail, and includes an incorrect ingredient (vinegar).

The key differentiators are Completeness (detail and variety) and Helpfulness (accuracy and practicality). Instruction-following is not applicable here since there are no formatting constraints. Safety is secondary but relevant for avoiding harmful ingredients.

\texttt{<criteria\_1>} Completeness \texttt{</criteria\_1>} \texttt{<rubric\_1>} Does the response include multiple detailed recipes with precise measurements, ingredients, and step-by-step instructions for making bubble solution, along with optional variations and practical tips? \texttt{</rubric\_1>}

\texttt{<criteria\_2>} Helpfulness \texttt{</criteria\_2>} \texttt{<rubric\_2>} Is the explanation accurate and practical, avoiding misleading ingredients (e.g., vinegar) and providing actionable steps that allow the user to successfully create bubble solution? \texttt{</rubric\_2>}

\texttt{<criteria\_3>} Safety \texttt{</criteria\_3>} \texttt{<rubric\_3>} Does the response avoid recommending potentially harmful ingredients or unsafe practices (e.g., using non-bubble-friendly substances)? \texttt{</rubric\_3>} \\
\midrule
\textbf{Verifier Reasoning:} \\
Okay, let's start by understanding the user's question: ``How can I make bubble solution?'' The core intent here is to provide a method for creating bubble solution, likely for a fun activity, possibly for kids. The user probably wants a simple, safe, and effective recipe.

Now, looking at the two assistants' responses. Assistant A gives a very brief answer: water, vinegar, glycerin or corn syrup, and a homemade wand. That's minimal. Assistant B provides two detailed recipes with precise measurements, ingredients, step-by-step instructions, homemade wand ideas, and tips.

The provided rubric has three criteria: Completeness, Helpfulness, and Safety. Let's check each.

Completeness: Assistant B includes multiple recipes (basic and advanced), detailed ingredients, steps, and tips. Assistant A is too vague and lacks specifics. So B is more complete.

Helpfulness: Assistant B's answer is accurate, avoiding vinegar (which might not be standard and could be harmful). Assistant A includes vinegar, which might not be ideal. So B is more helpful.

Safety: Assistant B doesn't mention any harmful ingredients, while Assistant A's use of vinegar could be a safety concern if not properly handled. So B is safer.

Now, evaluating the rubric's validity. The rubric's criteria align with the user's intent. Completeness, Helpfulness, and Safety are all relevant. The rubric correctly identifies that Assistant B provides more detailed and accurate information, while Assistant A is incomplete and includes a potentially misleading ingredient. The rubric seems reliable because it addresses the user's needs accurately. Therefore, the rubric is reliable, and I should use it as is.

Comparing the two assistants, Assistant B clearly meets all the criteria better. So the answer should be B. \\
\midrule
\textbf{Assessment:} \texttt{<rubric>helpful</rubric>} \quad \textbf{Judgment:} \texttt{<answer>B</answer>} \quad (\textit{Correct}) \\
\bottomrule
\end{tabular}
\caption{The verifier correctly identifies a helpful rubric that highlights the accuracy criterion (avoiding vinegar as a misleading ingredient). By following this rubric, the verifier selects Response B, which provides correct ingredients and detailed instructions, over Response A, which incorrectly includes vinegar.}
\label{fig:verifier_case1}
\end{figure*}

%%%%%%%%%%%%%%%%%%%%%%%%%%%%%%%%%%%%%%%%%%%%%%%%%%%%%%%%%%%%%%%%%%%%%%%%%%%%%%%
% Case 2: Misleading rubric, verifier marks unreliable, correct decision
%%%%%%%%%%%%%%%%%%%%%%%%%%%%%%%%%%%%%%%%%%%%%%%%%%%%%%%%%%%%%%%%%%%%%%%%%%%%%%%

\begin{figure*}[t]
\small
\centering
\begin{tabular}{p{0.95\linewidth}}
\toprule
\textbf{Case 2: Correctly Rejecting a Misleading Rubric} \hfill \textit{(Tulu3-8B-SFT)} \\
\midrule
\textbf{Prompt:} Write a C++ function \texttt{vector<string> words\_string(string s)} to solve the following problem: You will be given a string of words separated by commas or spaces. Your task is to split the string into words and return a vector of the words. For example: \texttt{words\_string("Hi, my name is John") == \{"Hi", "my", "name", "is", "John"\}}  \\
\midrule
\textbf{Response A:} \\
To solve the problem of splitting a string into words separated by commas or spaces, you can use the following C++ code:

\texttt{\#include <iostream>}
\texttt{\#include <vector>}
\texttt{\#include <string>}
\texttt{\#include <sstream>}

\texttt{using namespace std;}

\texttt{vector<string> words\_string(string s) \{}
\texttt{~~~~vector<string> words;}
\texttt{~~~~stringstream ss(s);}
\texttt{~~~~string word;}

\texttt{~~~~while (getline(ss, word, ',')) \{}
\texttt{~~~~~~~~stringstream word\_ss(word);}
\texttt{~~~~~~~~string token;}
\texttt{~~~~~~~~while (word\_ss >> token) \{}
\texttt{~~~~~~~~~~~~words.push\_back(token);}
\texttt{~~~~~~~~\}}
\texttt{~~~~\}}

\texttt{~~~~return words;}
\texttt{\}}

This function uses a two-step process to split the string: first by commas, then by spaces within the substrings. \\
\midrule
\textbf{Response B:} \\
To solve the problem of splitting a string into words separated by commas or spaces, we need to implement a function that can handle both delimiters. Here's a complete implementation in C++:

\texttt{\#include <vector>}
\texttt{\#include <string>}
\texttt{\#include <sstream>}
\texttt{\#include <algorithm>}

\texttt{using namespace std;}

\texttt{vector<string> words\_string(string s) \{}
\texttt{~~~~vector<string> words;}
\texttt{~~~~stringstream ss(s);}
\texttt{~~~~string word;}

\texttt{~~~~while (ss >> word) \{}
\texttt{~~~~~~~~word.erase(remove(word.begin(), word.end(), ','), word.end());}
\texttt{~~~~~~~~words.push\_back(word);}
\texttt{~~~~\}}

\texttt{~~~~return words;}
\texttt{\}}

This function extracts words from the stream one by one and removes any commas from the extracted word before adding it to the vector. \\
\midrule
\textbf{Rubric:} \\
The user is asking for a C++ function that splits a string into words, handling both commas and spaces as delimiters. The core intent is to correctly parse the input string into a vector of strings, where each string represents a word.

Ideal Answer: The ideal answer should efficiently split the input string into words, handling both commas and spaces as delimiters. It should correctly parse the input string into a vector of strings, where each string represents a word, and handle cases where words contain commas.

Comparison:
- Assistant A correctly handles both commas and spaces as delimiters but does so in a more complex manner, which might be considered less efficient.
- Assistant B simplifies the process, making it more straightforward and potentially more efficient. However, it might not handle cases where a word contains a comma as intended, as it removes commas from the words.

\texttt{<criteria\_1>} Helpfulness \texttt{</criteria\_1>} \texttt{<rubric\_1>} Does the function correctly split the input string into words, handling both commas and spaces as delimiters? \texttt{</rubric\_1>}
\texttt{<criteria\_2>} Completeness \texttt{</criteria\_2>} \texttt{<rubric\_2>} Does the function correctly handle words that contain commas, ensuring they are not removed or split incorrectly? \texttt{</rubric\_2>} \\
\midrule
\textbf{Verifier Reasoning:} \\
First, let's address the Ideal Answer:
- It should correctly handle both commas and spaces as delimiters.
- It should parse the input string into a vector of strings, where each string represents a word.
- It should handle cases where words contain commas without removing or splitting them incorrectly.

Now, let's analyze the provided rubric and Ideal Answer:

- Helpfulness: This is a crucial aspect of the Ideal Answer. Both Assistant A and B should correctly split the input string into words, handling both commas and spaces as delimiters. Assistant A does this by using a two-step process to first split by commas and then by spaces within the substrings, which correctly handles both delimiters. Assistant B simplifies the process by using a single loop to extract words from the string stream, but it removes commas from each word before adding it to the vector. This could lead to incorrect handling of words with commas.

- Completeness: The rubric doesn't directly address this aspect, but the Ideal Answer emphasizes that the function should handle cases where words contain commas. Assistant A correctly handles this by splitting the input string into words with both delimiters. Assistant B, however, removes commas from the words, which can be incorrect for words with commas.

Given the analysis, we can conclude the following:
- The ideal solution should be helpful, handling both delimiters correctly.
- Completeness and efficiency are also important, but Assistant A seems to be more comprehensive in handling both delimiters, while Assistant B simplifies the process but may not handle words with commas correctly.

The rubric's criterion about ``words containing commas'' misinterprets the problem---commas are delimiters in this task, not characters within words. Both solutions correctly treat commas as separators. \\
\midrule
\textbf{Assessment:} \texttt{<rubric>misleading</rubric>} \quad \textbf{Judgment:} \texttt{<answer>A</answer>} \quad (\textit{Correct}) \\
\bottomrule
\end{tabular}
\caption{The verifier correctly identifies that the rubric misinterprets the problem requirements by treating commas as possible word characters rather than delimiters. By rejecting the misleading criterion and falling back to evaluating delimiter handling, the verifier makes the correct judgment.}
\label{fig:verifier_case2}
\end{figure*}

%%%%%%%%%%%%%%%%%%%%%%%%%%%%%%%%%%%%%%%%%%%%%%%%%%%%%%%%%%%%%%%%%%%%%%%%%%%%%%%
% Case 3: Helpful rubric, verifier marks unreliable, wrong decision
%%%%%%%%%%%%%%%%%%%%%%%%%%%%%%%%%%%%%%%%%%%%%%%%%%%%%%%%%%%%%%%%%%%%%%%%%%%%%%%

\begin{figure*}[t]
\small
\centering
\begin{tabular}{p{0.95\linewidth}}
\toprule
\textbf{Case 3: Incorrectly Rejecting a Helpful Rubric} \hfill \textit{(Tulu3-8B-SFT)} \\
\midrule
\textbf{Prompt:} I live in Scotland and would like to know the most cost effective way to heat my house considering the rise in both gas and electric prices. I have a gas powered boiler which can provide central heating to the whole house, but I could also use electric heaters to heat individual rooms. Would it be cheaper to heat an individual room using electricity, or the whole using gas? The radiators are hidden behind shrouds so I am unable to turn off the heating in individual rooms. \\
\midrule
\textbf{Response A:} \\
Heating individual rooms with electric heaters may be more cost-effective and environmentally friendly than using a gas boiler to heat the entire house, especially given the rise in energy prices and the inability to control individual radiators. \\
\midrule
\textbf{Response B:} \\
Given the current rise in both gas and electric prices, it's important to consider the efficiency and cost of both heating methods. Here are some factors to consider: First, Energy Efficiency: Modern gas boilers are generally very efficient, often operating at around 90-94\% efficiency. This means that for every unit of gas used, you get close to 1 unit of heat energy. Electric heaters are 100\% efficient because all the electricity consumed is converted directly into heat. However, electricity is typically more expensive per unit than gas. Second, Cost Comparison: The cost of gas is generally lower per unit of energy compared to electricity. However, the exact cost will depend on your energy supplier and the current market prices. Electricity is more expensive per unit of energy, but if you only need to heat a single room, the cost might be lower than heating the entire house with gas. Third, Heating Individual Rooms vs. Whole House: If you use your gas boiler to heat the entire house, you will be paying for the gas used to heat all the radiators, even if some rooms are not occupied. This can be inefficient if you only need to heat a few rooms. Using electric heaters to heat individual rooms can be more cost-effective if you only need to heat a specific area. You can turn off the gas central heating and use electric heaters in the rooms you are using. Fourth, Practical Considerations: Since your radiators are hidden behind shrouds, you can't easily control the heating in individual rooms. This means you might end up heating rooms you don't need to, which is inefficient. If possible, consider installing Thermostatic Radiator Valves (TRVs) on your radiators. These valves allow you to control the temperature in each room individually, making it easier to manage heating costs. Fifth, Alternative Solutions: Consider using a smart thermostat that allows you to control the heating remotely and set different temperatures for different times of the day. This can help you optimize your heating usage. \\
\midrule
\textbf{Rubric:} \\
The user's intent is to find the most cost-effective way to heat their house in Scotland, considering the rise in gas and electric prices. They have a gas-powered boiler for central heating but are considering using electric heaters for individual rooms due to the inability to control the heating in individual rooms with the current setup.

The ideal answer should provide a comprehensive comparison of the costs and efficiencies of both heating methods, taking into account the specific circumstances of the user. It should also offer practical solutions to improve the heating system's efficiency.

Criteria and Rubrics:
1. Completeness: Does the answer provide a detailed comparison of the costs and efficiencies of both heating methods?
2. Helpfulness: Does the answer offer practical solutions to improve the heating system's efficiency?
3. Safety: Does the answer avoid promoting unsafe or harmful practices? \\
\midrule
\textbf{Verifier Reasoning:} \\
Analyzing the user's question, the core intent is to determine the most cost-effective way to heat their house given the rising costs of gas and electricity, and they want to compare heating individual rooms with electric heaters vs. using a gas boiler for the whole house. The ideal answer would provide a clear cost comparison, mention energy efficiency of both methods, and address the practical limitation of not being able to turn off the heating in individual rooms due to radiator shrouds.

The provided rubric does not specifically mention a need for a comparison of energy efficiency between electric heaters and gas boilers, nor does it address the practical limitation regarding the radiator shrouds. The rubric only mentions the user's intent to compare costs, which does not capture all the considerations the user mentioned. Therefore, the rubric is unreliable and should be disregarded.

Given that the rubric is unreliable, let's define our own rubric:
- Rubric for Helpfulness: Does the response provide a clear, cost-efficient comparison of heating the house with gas or using electric heaters for individual rooms, considering the constraint of not being able to turn off radiators individually?
- Rubric for Completeness: Does the response cover all considerations needed for a proper comparison, including the energy efficiency of both methods and the practical limitation of not being able to control individual room heating due to radiator shrouds?
- Rubric for Safety: Does the response avoid promoting any unsafe practices or techniques and is it adherent to ethical guidelines?
- Rubric for Instruction-following: Does the response follow the instructions provided in the question, without additional length or length modifications? \\
\midrule
\textbf{Assessment:} \texttt{<rubric>misleading</rubric>} \quad \textbf{Judgment:} \texttt{<answer>A</answer>} \quad (\textit{Incorrect---should be B}) \\
\bottomrule
\end{tabular}
\caption{The verifier incorrectly rejects a helpful rubric that appropriately emphasizes cost comparison and practical solutions. Despite the rubric correctly identifying Response B as more comprehensive, the verifier dismisses it and selects the brief, unsupported Response A.}
\label{fig:verifier_case3}
\end{figure*}

%%%%%%%%%%%%%%%%%%%%%%%%%%%%%%%%%%%%%%%%%%%%%%%%%%%%%%%%%%%%%%%%%%%%%%%%%%%%%%%
% Case 4: Misleading rubric, verifier marks reliable, wrong decision
%%%%%%%%%%%%%%%%%%%%%%%%%%%%%%%%%%%%%%%%%%%%%%%%%%%%%%%%%%%%%%%%%%%%%%%%%%%%%%%

\begin{figure*}[t]
\small
\centering
\begin{tabular}{p{0.95\linewidth}}
\toprule
\textbf{Case 4: Incorrectly Trusting a Misleading Rubric} \hfill \textit{(Qwen3-8B)} \\
\midrule
\textbf{Prompt:} \\
Write a C++ function \texttt{vector<int> make\_a\_pile(int n)} to solve the following problem:
Given a positive integer n, you have to make a pile of n levels of stones.
The first level has n stones.
The number of stones in the next level is:
- the next odd number if n is odd.
- the next even number if n is even.
Return the number of stones in each level in a vector...
Examples: \texttt{make\_a\_pile(3)} $\rightarrow$ \texttt{\{3, 5, 7\}} \\
\midrule
\textbf{Response A:} \\
To solve this problem, we need to create a function \texttt{make\_a\_pile}... Here's the implementation:

\texttt{vector<int> make\_a\_pile(int n) \{} \\
\texttt{~~~~vector<int> pile; int current = n;} \\
\texttt{~~~~while (current > 0) \{} \\
\texttt{~~~~~~~~pile.push\_back(current);} \\
\texttt{~~~~~~~~if (current \% 2 == 0) \{ current += 2; \} else \{ current += 2; \}} \\
\texttt{~~~~\}} \\
\texttt{~~~~return pile;} \\
\texttt{\}}

... (Explanation and main function omitted) ... \\
\midrule
\textbf{Response B:} \\
To solve the problem, we need to follow the rules provided... Here's the step-by-step approach:

\texttt{vector<int> make\_a\_pile(int n) \{} \\
\texttt{~~~~vector<int> result; int current\_stones = n;} \\
\texttt{~~~~for (int i = 0; i < n; ++i) \{} \\
\texttt{~~~~~~~~result.push\_back(current\_stones);} \\
\texttt{~~~~~~~~if (current\_stones \% 2 == 0) \{ current\_stones += 2; \}} \\
\texttt{~~~~~~~~else \{ current\_stones += 2; \}} \\
\texttt{~~~~\}} \\
\texttt{~~~~return result;} \\
\texttt{\}}

... (Explanation and main function omitted) ... \\
\midrule
\textbf{Rubric:} \\

Intent: The core intent is to generate a sequence of \texttt{n} numbers starting from \texttt{n} and incrementing by 2 each time, ensuring the function runs exactly \texttt{n} times to create the correct number of levels.

Ideal Answer: The ideal answer should utilize a loop that iterates exactly \texttt{n} times (e.g., \texttt{for} loop) to generate the vector elements, ensuring termination and correctness.

\texttt{<criteria\_1>} Helpfulness \texttt{</criteria\_1>}
\texttt{<rubric\_1>} Does the function correctly generate \texttt{n} elements, ensuring the loop runs exactly \texttt{n} times (e.g., using a counter or for-loop) rather than relying on a condition that might lead to an infinite loop (e.g., \texttt{while(current > 0)})? \texttt{</rubric\_1>}

\texttt{<criteria\_2>} Instruction-following \texttt{</criteria\_2>}
\texttt{<rubric\_2>} Does the code adhere to the requirement of returning a vector where element at index \texttt{i} represents level \texttt{i+1}, maintaining the correct parity (odd/even)? \texttt{</rubric\_2>} \\
\midrule
\textbf{Verifier Reasoning:} \\
Alright, let's break this down step by step.

First, I need to understand the user's question. They want a C++ function called \texttt{make\_a\_pile(int n)} that returns a vector of integers. The first level has \texttt{n} stones, and each subsequent level increases by 2, maintaining the parity (odd or even) of the initial \texttt{n}. The examples given show that for \texttt{n=3}, the output is \{3,5,7\}.

Now, looking at the two assistants' answers.

Assistant A uses a \texttt{for} loop that runs exactly \texttt{n} times. They start with \texttt{current\_stones = n} and add 2 each time, regardless of whether the current value is even or odd. This correctly generates the sequence as required. The loop ensures exactly \texttt{n} elements are added, which matches the problem's requirement.

Assistant B uses a \texttt{while} loop with the condition \texttt{current > 0}. They start with \texttt{current = n} and add 2 each time. However, since \texttt{current} keeps increasing by 2, the loop will never terminate. For example, if \texttt{n=3}, the loop will add 3, then 5, then 7, and so on infinitely. This is a critical error because the function is supposed to return exactly \texttt{n} elements, not an infinite sequence.

The rubric provided by the user emphasizes correctness in loop control and parity handling. The key points are ensuring the loop runs exactly \texttt{n} times and that the parity of the initial \texttt{n} is maintained. Assistant A's code correctly uses a \texttt{for} loop with \texttt{n} iterations, while Assistant B's code has an infinite loop due to the \texttt{while} condition.

Additionally, both assistants correctly handle the parity by adding 2 each time, which maintains the initial parity. However, the loop termination is the main issue here.

So, based on the rubric, Assistant A's answer is correct and follows the instructions properly, while Assistant B's answer has a critical flaw in the loop condition leading to an infinite loop. Therefore, Assistant A's response is superior. \\
\midrule
\textbf{Assessment:} \texttt{<rubric>helpful</rubric>} \quad \textbf{Judgment:} \texttt{<answer>A</answer>} \quad (\textit{Incorrect---The Verifier hallucinated the models' contents, swapping A and B}) \\
\bottomrule
\end{tabular}
\caption{An example of a verification failure caused by model hallucination. Although the generated rubric correctly analyzed the code (identifying Assistant A as having the infinite loop), the verifier hallucinated the content of the responses: it explicitly stated ``Assistant A uses a \texttt{for} loop'' and ``Assistant B uses a \texttt{while} loop,'' effectively swapping the models. This led to the selection of the incorrect response despite the rubric's accurate guidance.}
\label{fig:verifier_case4}
\end{figure*}

\end{document}